\documentclass[10pt,twocolumn,letterpaper]{article}
\usepackage[pagenumbers]{cvpr} 
\usepackage[accsupp]{axessibility}

\usepackage[most]{tcolorbox}
\tcbuselibrary{skins,breakable}
\tcbuselibrary{listingsutf8}
\tcbuselibrary{external}
\usepackage{graphicx}
\usepackage{lipsum} 
\usepackage{listings}

\usepackage[ruled]{algorithm2e}
\SetArgSty{textnormal}
\SetAlgoNoEnd

\definecolor{cvprblue}{rgb}{0.21,0.49,0.74}
\usepackage[pagebackref,breaklinks,colorlinks,allcolors=cvprblue]{hyperref}

\definecolor{myyellow}{RGB}{255, 250, 205}
\definecolor{myblue}{RGB}{68, 114, 196}
\definecolor{myred}{RGB}{220, 20, 60}
\definecolor{mygreen}{RGB}{46, 139, 87}
\definecolor{myorange}{RGB}{237, 125, 49}
\definecolor{dpurple}{RGB}{128, 0, 128}
\definecolor{lpurple}{RGB}{235, 232, 242}
\definecolor{lblue}{rgb}{0.94, 0.96, 1.0}
\definecolor{grey}{RGB}{128, 128, 128}
\definecolor{lred}{RGB}{255,114,118}
\definecolor{mypurple}{RGB}{75,0,130}
\hypersetup{
    colorlinks=true,
    linkcolor=dpurple,
    filecolor=cvprblue,      
    urlcolor=cvprblue,
    citecolor=cvprblue,
}
\usepackage{comment}
\usepackage{multirow}

\usepackage{xspace}
\usepackage{wrapfig}
\usepackage{amsfonts}
\usepackage{bbm}

\newcommand{\mypar}[1]{\vspace{1mm}\noindent\textbf{#1}}
\newcommand{\myparit}[1]{\vspace{1mm}\noindent\emph{#1}}

\def\method{VADAR\xspace}
\def\ourbench{\textsc{Omni3D-Bench}\xspace}
\def\clevr{\textsc{CLEVR}\xspace}

\def\gpt4o{GPT4o\xspace}
\def\llama{Llama3.2\xspace}
\def\gemini{Gemini\xspace}
\def\claude{Claude-Sonnet\xspace}
\def\viper{ViperGPT\xspace}
\def\visprog{VisProg\xspace}
\def\leftm{LEFT\xspace}

\title{Visual Agentic AI for Spatial Reasoning with a Dynamic API}
\author{%
  Damiano Marsili$^{*}$ \quad 
  Rohun Agrawal$^{*}$ \quad 
  Yisong Yue \quad
  Georgia Gkioxari \smallskip \smallskip\\
  California Institute of Technology \\
}

\begin{document}

\twocolumn[{%
\renewcommand\twocolumn[1][]{#1}%
\maketitle
\begin{center}
    \vspace{-10mm}
    \centering
    \includegraphics[width=0.94\linewidth]{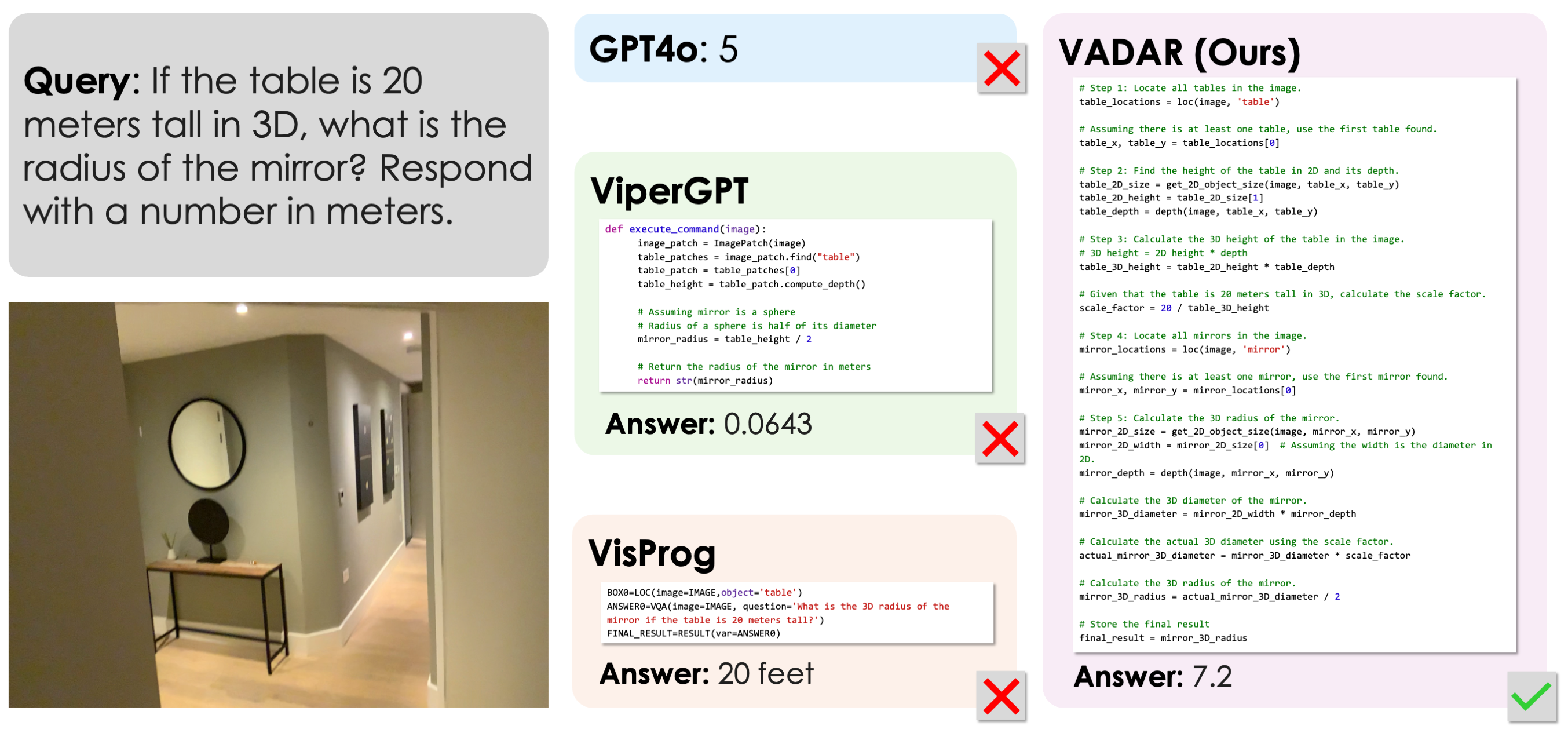}
    \vspace{-4mm}
    \captionof{figure}{Spatial reasoning in 3D is challenging as it requires multiple steps of grounding and inference. We introduce a benchmark for 3D understanding with complex queries; an example is shown here. To tackle these queries we propose a training-free agentic approach, \method, that dynamically generates new skills in Python and thus can handle a wider range of queries compared to prior methods.} 
    \label{fig:fig1}
\end{center}%
}]
\begin{abstract}
Visual reasoning -- the ability to interpret the visual world -- is crucial for embodied agents that operate within three-dimensional scenes. 
Progress in AI has led to vision and language models capable of answering questions from images. However, their performance declines when tasked with 3D spatial reasoning. To tackle the complexity of such reasoning problems, we introduce an agentic program synthesis approach where LLM agents collaboratively generate a Pythonic API with new functions to solve common subproblems. Our method overcomes limitations of prior approaches that rely on a static, human-defined API, allowing it to handle a wider range of queries. To assess AI capabilities for 3D understanding, we introduce a new benchmark of queries involving multiple steps of grounding and inference.
We show that our method outperforms prior zero-shot models for visual reasoning in 3D and empirically validate the effectiveness of our agentic framework for 3D spatial reasoning tasks. Project website: \href{https://glab-caltech.github.io/vadar/}{https://glab-caltech.github.io/vadar/}
\end{abstract}
\vspace{-4mm}
\begin{minipage}{\textwidth}
\makeatletter\def\Hy@Warning#1{}\makeatother
\footnotetext{$^*$Equal contribution.}
\end{minipage}
\section{Introduction}
\label{sec:intro}
Consider \cref{fig:fig1}. 
Here, a person or an agent wants to determine the radius of the mirror in the image, given that the table is 20 meters tall. 
Answering this question requires visual reasoning, a crucial step toward achieving general-purpose AI.
Visual reasoning enables machines to analyze and make sense of the visual world. 
Humans rely heavily on visual cues to navigate complex environments, interact with objects and make informed decisions.
Our goal is to build intelligent agents that can do the same.
Recent advances in AI have produced vision and language models (VLMs)~\cite{gpt4, claude, molmo, gemini} that can answer questions from images. 
Although impressive, these models excel primarily at category-level semantic understanding. 
Their performance significantly declines when tasked with spatial understanding within the three-dimensional world~\cite{spatial, cambrian1, whatsup}. 

Returning to \cref{fig:fig1}, to answer the query, an AI agent must first locate the relevant objects,  determine their dimensions in pixel space, use their depth to calculate their 3D sizes, and finally compute the mirror's radius using the table's height.
This is a complex sequence of tasks, involving multiple steps of understanding, grounding, and inference. 
\gpt4o~\cite{gpt4}, a state-of-the-art VLM trained on extensive datasets, gives a wrong final answer.

To address the complexity of 3D spatial reasoning tasks, we propose a system of agents working collaboratively to create executable programs for a given image. 
Our approach leverages LLM agents that \emph{dynamically} define and expand a domain-specific language (DSL) \emph{as needed}, generating new functions, skills and reasoning, in two phases: the \textbf{API Generation} and the \textbf{Program Synthesis} stage.
Vision specialists -- an object detector, a depth estimator and object attribute predictor -- help the agents execute the program.
We name our approach \method, as it integrates Visual, Agentic, Dynamic AI for Reasoning.
\method belongs in the family of visual program synthesis methods, like \viper~\cite{vipergpt} and \visprog~\cite{visprog}, but addresses a key limitation in these approaches: their reliance on a static, human-defined DSL, which restricts them to a predefined range of functionality.
This limitation is evident in~\cref{fig:fig1}, where \viper generates an incomplete, inaccurate program and \visprog defaults to a holistic visual question answer (VQA) approach for answering the query.
\method's output in~\cref{fig:fig1} demonstrates its ability to tackle a wider range of visual queries.

We evaluate 3D spatial reasoning using challenging benchmarks designed for rigorous assessment of 3D understanding. Our evaluation includes \clevr~\cite{clevr} and our newly introduced benchmark, \ourbench, based on Omni3D~\cite{omni3d}; \cref{fig:fig1} shows an example. Both datasets emphasize visual queries involving relative depth, size, and object location, often conditioned on measurement hypotheses, requiring grounding and 3D inference. This contrasts with previous spatial reasoning benchmarks like GQA~\cite{gqa}, which primarily emphasize appearance-based reasoning.

At a high level, \method roughly mirrors the workflow of a software engineer when defining, implementing, and testing new software solutions for a given problem. Leveraging its agentic design, \method autonomously defines and implements functions such as \texttt{\_find\_closest\_object\_3D}, \texttt{\_is\_behind}, \texttt{\_count\_objects\_by\_attributes\_and\_position}, \texttt{\_is\_left\_of}, and more. These functions are used by the Program Agent, resulting in more concise programs, less output tokens and thus a lower likelihood of errors from LLM-generated predictions.
We empirically show that \method outperforms a \emph{no-API} agent by 6\%, highlighting the value of general, reusable, functions within an API.
Moreover, we show that our generated API significantly surpasses a static, human-defined API used in~\cite{visprog, vipergpt}, by more than 20\% on CLEVR.
\method performs competitively with state-of-the-art VLMs, on \ourbench, while also providing executable programs. 

Considering the rapid progress in AI, one might wonder if methods like \method can dominate monolithic VLMs in 3D spatial reasoning. One clear advantage of \method is its ability to generate interpretable programs. However, our experiments highlight another key potential. Improving VLMs for 3D reasoning would require extensive datasets of image-question-answer tuples with 3D information, an onerous endeavor. In contrast, our experiments show that if the component vision models -- an object detector, an attribute predictor and depth estimator -- were replaced with oracle versions, \method would achieve 83.0\% accuracy, 24\% higher from the best VLM.
This indicates that \method is bottlenecked by the performance of its vision specialists. Thus, an alternative path to scaling 3D spatial reasoning could be through improving specialized vision models, which tackle a simpler problem than general-purpose VQA and for which training data is more readily available.

\section{Related Work}
\label{sec:related}

Our work draws from areas of language modeling, visual program synthesis and library learning. 

\mypar{VLMs for Spatial Reasoning.}
LLMs~\cite{gpt4,gemini,llama3,claude} are trained on large corpora of text, including domain specific languages (DSLs) such as Python. Their multi-modal variants incorporate images and are additionally trained on image-text pairs showing impressive results for visual captioning and vision question-answering (VQA)~\cite{vqa}. 
Despite their strong performance, their ability to reason beyond category-level semantic queries is limited.
Recent work~\cite{cambrian1,whatsup} shows that VLMs suffer on visual tasks such as grounding spatial relationships and inferring object-centric attributes. SpatialRGPT~\cite{spatialrgpt} and SpatialVLM~\cite{spatial} use data synthesis pipelines to generate templated queries for spatial understanding. We compare to SpatialVLM and show that it struggles to tackle 3D spatial reasoning queries.

\mypar{Visual Program Synthesis.}
Recent advances in visual reasoning have led to methods which improve upon the capabilities of vision-based models by composing them symbolically via program synthesis. 
\visprog~\cite{visprog} prompts an LLM to generate an executable program of a specified DSL that calls and combines vision specialists -- OwlViT~\cite{owlvit} for object detection, CLIP~\cite{clip} for classification, and ViLT~\cite{vilt} for VQA.
\viper~\cite{vipergpt} directly generates Python code by providing a Python API specification to the LLM agent and adds MiDaS~\cite{midas} as the vision specialist for depth estimation, in addition to GLIP~\cite{glip} and X-VLM~\cite{x-vlm} for vision-language tasks.
Both approaches rely on a pre-defined DSL, which narrows the scope of applicability and makes these methods difficult to extend to a wider range of queries.
Similar to \viper, we use Python as the interface for our LLM agents, but we don't define the API a-priori.
We instead rely on our agentic workflow to generate the API needed to tackle complex spatial reasoning queries.
We compare to \viper and \visprog and show that both struggle to generate accurate programs for complex queries, often completely ignoring part of the query. 

\mypar{Library Learning.}
An emerging field in LLM research focuses on the dynamic creation and extension of a set of reusable functions during problem-solving. 
Early work on library learning predates the use of LLMs \cite{ellis2023dreamcoder,valkov2018houdini,lake2015human},
and focuses on a common architecture of iteratively proposing new programs and synthesizing commonly used components into a library.
Modern approaches follow this same paradigm, but use LLMs to accelerate the synthesis of useful programs, applied to gaming \cite{wang2023voyager}, 3D graphics scripting \cite{hu2024scenecraft}, theorem proving \cite{thakur2024context}, and symbolic regression \cite{grayeli2024symbolic}. 

\mypar{Neuro-symbolic AI} generates interpretable symbolic components for complex tasks 
and has been explored for a wide range of fields, including spatial reasoning~\cite{mao2019neuro}, grounding of 3D point clouds~\cite{ns3d}, mechanistic modeling in scientific domains \cite{grayeli2024symbolic,sun2022neurosymbolic}, logical reasoning \cite{olausson2023linc}, amongst other areas. 
Closer to us is the logic-enhanced LLM, \leftm~\cite{whatsleft}, that uses a dynamic DSL of first order logic structures and differentiably executes them using domain-specific modules. These modules, instantiated as MLPs, ground spatial concepts, \eg \emph{``is left of"}, and are \emph{trained with supervision}.
On \clevr, \method, which is \emph{training-free}, achieves the same performance as \leftm when trained with $\geq 10,000$ training samples. 
A benefit of our training-free approach is that it scales to new domains where 3D supervision is hard to acquire, as we show on our \ourbench.

\mypar{Spatial Reasoning Benchmarks.}
Existing benchmarks test aspects of visual reasoning with free-form language~\cite{laurent2024lab, arc}.
We focus on natural-image based ones. 
VQA~\cite{vqa} introduced the task of visual question answering.
GQA~\cite{gqa} is a popular large-scale VQA benchmark with questions that pertain to object and attribute recognition, of mostly a single-step inference -- \emph{``What color is the cat next to the chair?"}, \emph{``What type of vehicle is on top of the road?"}, \emph{``Do the wildflowers look ugly?"}. 
RefCOCO~\cite{refcoco} targets object localization with referring expressions such as \emph{``the man in a red shirt"}. 
What's up~\cite{whatsup} quantifies comprehension of basic 2D spatial relations such as \emph{``left of"} and \emph{``above"}.
These benchmarks evaluate aspects of visual reasoning, but critically omit 3D understanding. 
Q-Spatial Bench~\cite{qspatial} focuses solely on absolute 3D measurements.
Cambrian-1~\cite{cambrian1} proposes a VQA benchmark repurposing images and annotations from Omni3D~\cite{omni3d}, but its queries focus on the relative depth and depth ordering of objects with (2 or 3)-choice questions.
Our benchmark also repurposes Omni3D annotations, but in contrast to Cambrian-1, we design more complex queries that extend beyond depth ordering and multiple choice. Concurrent to our work, VSI-Bench~\cite{thinkinginspace} introduces a video understanding benchmark focused on spatial relationships, which we discuss extensively in Appendix \ref{sec:vsi-bench}.

\section{Method}
\label{sec:method}

\begin{figure*}[t!]
  \centering
  \includegraphics[width=1.0\linewidth]{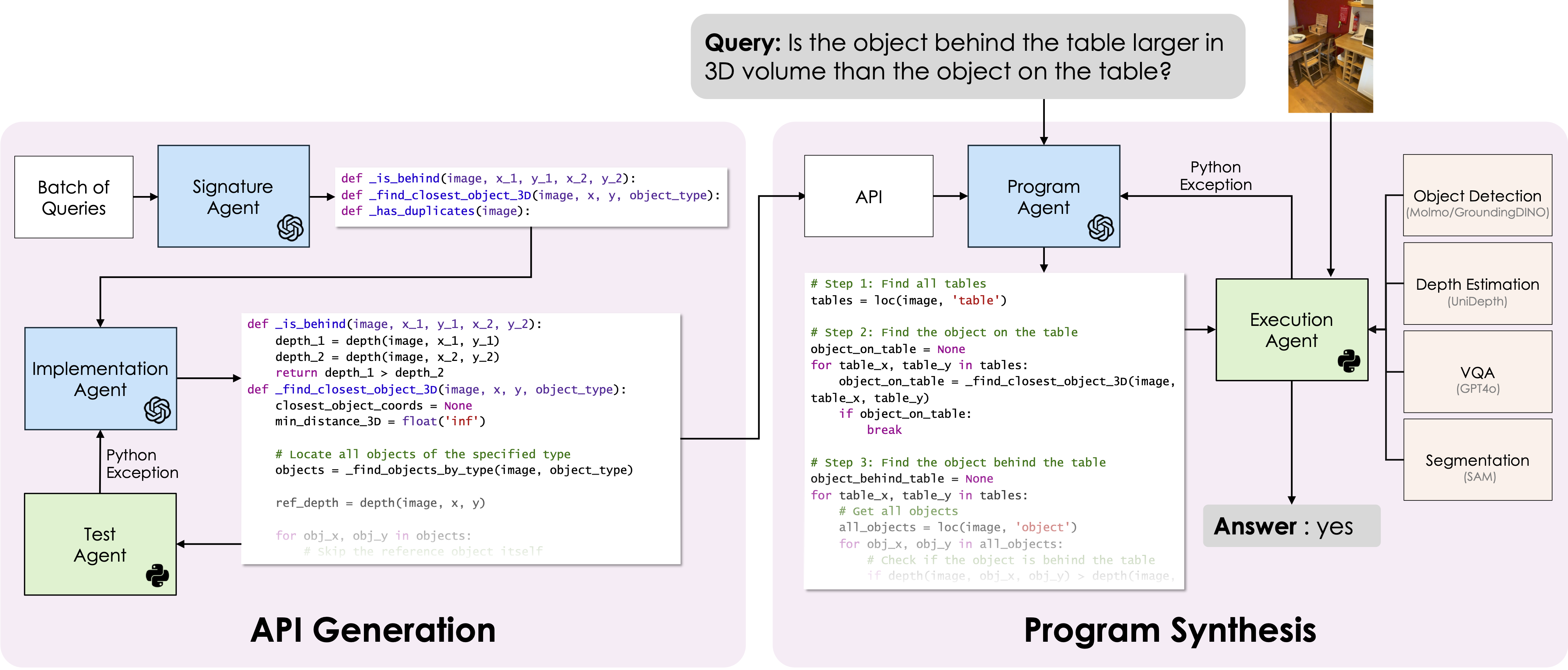}
  \vspace{-7mm}
  \caption{\textbf{Overview.} \method consists of an API generation stage and a program synthesis stage. The Signature \& Implementation Agents generate an API that is used by the Program Agent to produce a program to answer the question, executed by the Execution Agent.}
  \label{fig:overview}
  \vspace{-3mm}
\end{figure*}

At the core of our approach is a dynamic API generated by LLMs that can be extended to address new queries that require novel skills. The goal of the API is to break down complex reasoning problems into simpler subproblems with general modules that can be used during program synthesis. Our approach consists of an API Generation stage and a Program Synthesis stage, illustrated in \cref{fig:overview}.

\myparit{Vision Specialists.}
During program execution on the image, we employ vision models for solving visual subtasks:
Molmo's~\cite{molmo} pointing model and GroundingDINO~\cite{groundingdino} are used to localize objects prompted with text ($\texttt{loc}$), SAM~\cite{sam} returns the bounding box from the object's mask prompted with Molmo's points ($\texttt{get\_2D\_object\_size}$), UniDepth~\cite{unidepth} estimates the depth at an image location ($\texttt{depth}$), GPT4o is utilized as a VQA module to query object attributes (color, material) from an image with the target object bounding box overlayed ($\texttt{vqa}$). We initialize the API with these functions. The API also includes $\texttt{same\_object}$ that computes the overlap of two object bounding boxes to determine if the objects are the same.

\subsection{API Generation}
\vspace{-2mm}
\SetAlgoLined
\DontPrintSemicolon
\begin{algorithm}[]
\caption{\method: API Generation}
\label{algo:api}
\SetKwData{Data}{Data}
\SetKw{Continue}{continue}
\SetKwInOut{KwInit}{Initialize}
\SetKwProg{Fn}{Function}{}{end}

\KwData{Questions $\mathcal{Q}$}
$\mathcal{S} \gets \{\} $\tcp*[r]{Signatures}
$\mathcal{A} \gets \{\text{Vision Models}\}$\tcp*[r]{API Methods}
\For{batch $B \subset \mathcal{Q}$} {
    $\mathcal{S} \gets \mathcal{S} \cup  \texttt{SignatureAgent}(B)$
}
\For{$S \in \mathcal{S}$} {
    $e_S \gets 0$ \tcp*[r]{Error count}
    $A \gets \texttt{ImplementationAgent}(S)$ \;
    $E \gets \texttt{TestAgent}(A)$\;
    \eIf{Python Exception $E$}{
        \lIf{$e_S = 5$}{\Continue}
        \uElseIf{$E$ is ``undefined method $U$''}{
            $e_S \gets e_S + 1$\;
            Recursively implement $U$
        }
        \Else{
            $e_S \gets e_S + 1$\;
            Re-implement $S$ using $E$
        }
    } {
    $\mathcal{A} \gets \mathcal{A} \cup A$
    }
}
\Return{$\mathcal{A}$}
\end{algorithm}
\vspace{-3mm}
\cref{algo:api} describes the API Generation. Here, the \textbf{Signature Agent} and the \textbf{Implementation Agent} collaborate to define and implement new functions \emph{as needed} to aid in solving the queries. 
First, the Signature Agent receives a batch of $N$ queries ($N=15$), \emph{without answers}, and is instructed to produce general method signatures for subproblems that could arise when answering those kinds of queries. The Implementation Agent then implements the signatures in Python.
Examples of signatures and their implementations are shown in \cref{fig:overview}.

\myparit{Prompting the Signature Agent.}
The agent receives the current API state as docstrings so it avoids duplicating existing methods. We observed that our Signature Agent performed better without in-context examples as it produced a more diverse API with wider potential functionality. 

\myparit{Prompting the Implementation Agent.}
The Implementation Agent receives all other signatures in the API along with the signature it needs to implement, so it can use other API methods in its implementation, enabling a hierarchy in the API. In contrast to the Signature Agent, providing in-context examples significantly enhances the Implementation Agent’s output, as implementation prioritizes accuracy over diversity. We refer to these examples as \emph{weak} in-context learning (ICL), as they guide correct method implementation in Python, unlike \emph{strong} ICL, which breaks down queries into full programs. Prompts for both agents and weak-ICL examples are found in the Appendix. 

\myparit{Depth-First Implementation}.
Once a method is implemented from its signature, the Test Agent, a Python interpreter, runs it using placeholder inputs. If a runtime error occurs, the Test Agent signals the Implementation Agent to revise it with the exception message. However, if the implementation relies on another yet-to-be-implemented API method, the test run cannot proceed. In this case, the Implementation Agent traverses an implicit dependency graph, depth-first, ensuring that prerequisite methods are implemented first (see Algo.~\ref{algo:api}).

Consider the following example where the signatures \texttt{get\_color}, \texttt{find\_objects\_by\_color}, \texttt{count\_objects\_left\_of}, and \texttt{is\_left\_of}, are defined by the Signature Agent, in that order. First, the Implementation Agent will implement \texttt{get\_color}, the Test Agent will be called, and barring  no runtime errors, the method will be complete. Then, the implementation for \texttt{find\_objects\_by\_color} uses \texttt{get\_color}, which is implemented, so the Test Agent only checks for Python errors. If \texttt{count\_objects\_left\_of} attempts to use \texttt{is\_left\_of}, the Test Agent will detect that \texttt{is\_left\_of} is not implemented and recursively call the Implementation Agent to implement \texttt{is\_left\_of}, followed by \texttt{count\_objects\_left\_of}.

In the event a cycle in the dependency graph is persistent after attempting the implementation of those methods 5 times, the methods in the cycle are deleted. Empirically, we rarely detect such cycles, which can be attributed to the Signature Agent producing multiple signatures at once, tending to avoid proposing signatures that overlap in function.

\subsection{Program Synthesis}
\begin{algorithm}[]
\caption{\method: Program Synthesis}
\label{alg:cap}
\SetKwData{Data}{Data}
\SetKw{Continue}{continue}
\SetKwInOut{KwInit}{Initialize}
\SetKwProg{Fn}{Function}{}{end}

\KwData{Image-Query pairs $\mathcal{D} = \{(I, Q)\},$ \; 
\qquad \space\space API methods $\mathcal{A}$ }
$\mathcal{R} \gets \{\}$\tcp*[r]{Results}
\For{$(I, Q) \in \mathcal{D}$} {
    $e_P \gets 0$ \tcp*[r]{Error count}
    $P \gets \texttt{ProgramAgent}(Q, \mathcal{A})$ \;
    $E, R \gets \texttt{ExecutionAgent}{(P, I, \text{Vision Models})}$ \;
    \eIf{Python Exception $E$ $\textbf{and } e_P < 5$}{
        $e_P \gets e_P + 1$ \;
        Re-generate $P$ using $E$\;
    }
    {
    $\mathcal{R} \gets \mathcal{R} \cup R$\;
    }
}
\Return{$\mathcal{R}$}\;
\end{algorithm}
 
The \textbf{Program Agent} receives the generated API and a single question as input. Its task is to generate Python code that leverages the API to solve the question. The Execution Agent, another Python interpreter, executes the program line-by-line. In the event of a Python error, it provides the Program Agent with the exception, and a new program is generated. This is repeated at most 5 times, after which the program returns an execution error.

\myparit{Prompting the Program Agent.} 
Following the success of Chain-of-Thought (CoT) prompting~\cite{cot}, we instruct the Program Agent to create a plan before generating the corresponding program. In-context examples boost the Program Agent’s performance. However, unlike \visprog~\cite{visprog} and \viper~\cite{vipergpt} that use strong-ICL, we use API-agnostic natural language instructions since the API is not predefined, making it impossible to provide full program examples. These instructions help for the same reason as with the Implementation Agent, to focus on correctness. The prompt for the Program Agent is provided in the Appendix.

\myparit{Test \& Execution Agent vs Critics.}
In modern library learning, LLM agents, or critics, evaluate the quality and utility of learned functions. Our Test and Execution Agents also assess method quality, but we opt for deterministic critics that leverage the full Python runtime, signaling LLM Agents with Python exceptions in case of errors.

\vspace{-1mm}
\section{Experiments}
\label{sec:experiments}\

We evaluate our approach on challenging spatial reasoning benchmarks, demonstrating that a dynamically generated API outperforms the static, human-defined APIs in ViperGPT~\cite{vipergpt} and VisProg~\cite{visprog} by a large margin. Additionally, we compare against state-of-the-art monolithic VLMs trained on billions of (image, question, answer) samples, showing that our method competes favorably and even surpasses them on certain question types while offering interpretable reasoning steps for complex queries.

\subsection{A Benchmark for Spatial Reasoning in 3D}
We evaluate 3D spatial reasoning using \clevr, and our newly introduced benchmark, \ourbench.

\mypar{\clevr}~\cite{clevr} consists of (image, question, answer) tuples. Each image contains 2-10 objects of 3 different shapes, 8 colors, 2 materials, and 2 sizes. Despite the simplicity of the scenes, the questions in \clevr are complex, \eg, \emph{``There is a large ball right of the large metal sphere that is left of the large object that is behind the small brown sphere; what color is it?"}. Our \clevr benchmark contains 1,155 samples, 400 of which require a numerical answer, 399 are yes/no questions, and 356 are multiple-choice questions.

\mypar{\ourbench} is sourced from Omni3D~\cite{omni3d}, a dataset of images from diverse real-world scenes with 3D object annotations. We repurpose images from Omni3D to a VQA benchmark, with questions about 3D information portrayed in the image, such as \emph{``If the height of the front most chair is 6 meters in 3D, what is the height in 3D of the table in the image?"} and \emph{``How many bottles would you have to stack on top of each other to make a structure as tall in 3D as the armchair?"}. \ourbench complements \clevr with \emph{non-templated} queries pertaining to 3D locations and sizes of objects. 
Our queries test 3D reasoning, as they require grounding objects in 3D and combining predicted attributes to reason about distances and dimensions in three dimensions. \ourbench consists of 500 extremely challenging (image, question, answer) tuples. 

We compare our proposed benchmark to GQA~\cite{gqa}, a popular visual reasoning dataset. GQA derives queries from scene graphs which primarily pertain to the visual appearance and attributes of objects. Example queries in GQA are \emph{``Is there a red truck or bus?"}, \emph{``Is the field short and brown?"} and \emph{``Is the chair in the top part of the image?"}. These are significantly simpler to queries in \clevr and \ourbench which involve multiple steps of grounding and inference in two- and three- dimensions.

\begin{table*}[]
\resizebox{0.99 \linewidth}{!}{
\begin{tabular}{ll|cccc|ccccc}
                                & & \multicolumn{4}{c|}{\clevr}         & \multicolumn{4}{c}{\ourbench}             \\
                                &   & numeric & y/n & multi-choice & Total & numeric (ct) & numeric (other) & y/n & multi-choice & Total \\
\hline
\multirow{6}{*}{\rotatebox[origin=c]{90}{\small VLMs}} 
                                & GPT4o~\cite{gpt4}                   & 52.3 & 63.0 & 60.0 & 58.4 & \bf{28.1} & \bf{35.5} & \bf{66.7} & 57.2 & \bf{42.9} \\
                                & Claude3.5-Sonnet~\cite{claude}      &  44.7 & 61.4 & \bf{72.2} & \bf{58.9} & 22.4 & 20.6 & 62.2 & 50.6 & 32.2 \\
                                & Llama3.2~\cite{llama3}              & 34.6 & 45.6 & 49.0 & 42.8 & 24.3 & 19.3 & 47.5 & 27.4 & 25.6 \\
                                & Gemini1.5-Pro~\cite{gemini}         & 44.9 & 59.7 & 67.0 & 56.9 & 25.2 & 28.1 & 46.2 & 37.6 & 32.0 \\
                                & Gemini1.5-Flash~\cite{gemini}       & 43.1 & 58.8 & 56.8 & 52.8 & 24.3 & 27.6 & 51.1 & 52.9 & 35.0 \\
                                & Molmo~\cite{molmo}                  & 11.0 & 42.6 & 51.4 & 34.4 & 21.4 & 21.7 & 29.3 & 41.2 & 26.1 \\
                                & SpaceMantis~\cite{mantis, spatial}  & 14.5 & 52.9 & 32.3 & 33.2 & 20.0 & 21.7 & 50.6 & 48.2 & 30.3 \\
\hline
\multirow{3}{*}{\rotatebox[origin=c]{90}{\small \shortstack{Program\\Synthesis}}} 
                                & ViperGPT~\cite{vipergpt}            & 20.5 & 43.4 & 13.4 & 26.2 & 20.0 & 15.4 & 56.0 & 42.4 & 26.7 \\
                                
                                & VisProg~\cite{visprog}              & 16.7 & 48.4 & 28.3 & 31.2 & \phantom{ } 2.9 & \phantom{ } 0.9 & 54.7 & 25.9 & 13.5 \\
                                & \method (ours)                      & \bf{53.3} & \bf{65.3} & 40.8 & 53.6 & 21.7 & \bf{35.5} & 56.0 & \bf{57.6} & 40.4 \\
\end{tabular}
}
\vspace{-2mm}
\caption{\textbf{Accuracy (\%) on \clevr and \ourbench.} We compare to state-of-the-art monolithic VLMs and Program Synthesis approaches. For each benchmark, we breakdown performance for \emph{numeric (ct)}, \emph{numeric (other)}, \emph{yes/no} and \emph{multiple-choice} answers and report total accuracy. For \emph{numeric (other)} queries, which require floating point answers, we report MRA. \method outpeforms \viper and \visprog with a big margin. \method outperforms all large VLMs on \ourbench except \gpt4o, which it is narrowly behind.}
\label{tab:main_results}
\vspace{-3mm}
\end{table*}

\begin{table*}[]
\resizebox{0.99 \linewidth}{!}{
\begin{tabular}{l|cccc|ccccc}
                                & \multicolumn{4}{c|}{\clevr}         & \multicolumn{4}{c}{\ourbench}             \\
                                & numeric & y/n & multi-choice & Total & numeric (ct) & numeric (other) & y/n & multi-choice & Total \\
\hline
                                ViperGPT~\cite{vipergpt}            & 38.5 & 57.8 & 30.2 & 42.6
                                & 50.0 & 17.8 & 66.7 & 49.3 & 54.9 \\
                                
                                VisProg~\cite{visprog}              & 25.3 & 52.5 & 41.8 & 39.9 & \bf{100.0} & 23.5 & 68.5 & 66.7 & 66.0 \\
                                \method (ours)                      & \bf{82.4} & \bf{85.4} & \bf{81.0} & \bf{83.0} & \bf{100.0} & \bf{82.3} & \bf{100.0} & \bf{94.1} & \bf{94.4} \\
                                \hline
                                GPT4o & 52.3 & 63.0 & 66.0 & 58.4 & 30.0 & 29.4 & 77.8 & 44.0 & 53.7 \\
                                Claude3.5-Sonnet & 44.7 & 61.4 & 72.2 & 58.9 & 30.0 & 35.3 & 83.3 & 56.0 & 59.3
                                
\end{tabular}
}
\vspace{-2mm}
\caption{\textbf{Oracle accuracy (\%) on \clevr and \ourbench.} We assess program synthesis correctness by replacing vision specialists with oracle variants. We report oracle accuracy on \clevr and a smaller subset of \ourbench and compare to best performing monolithic VLMs on the same sets. \method's high oracle accuracy indicates its main limitation is the vision specialists' performance.}
\label{tab:oracle_results}
\vspace{-3mm}
\end{table*}


\subsection{Results on Spatial Reasoning in 3D}

\begin{figure}[t!]
    \vspace{-5mm}
    \centering
    \includegraphics[width=\linewidth]{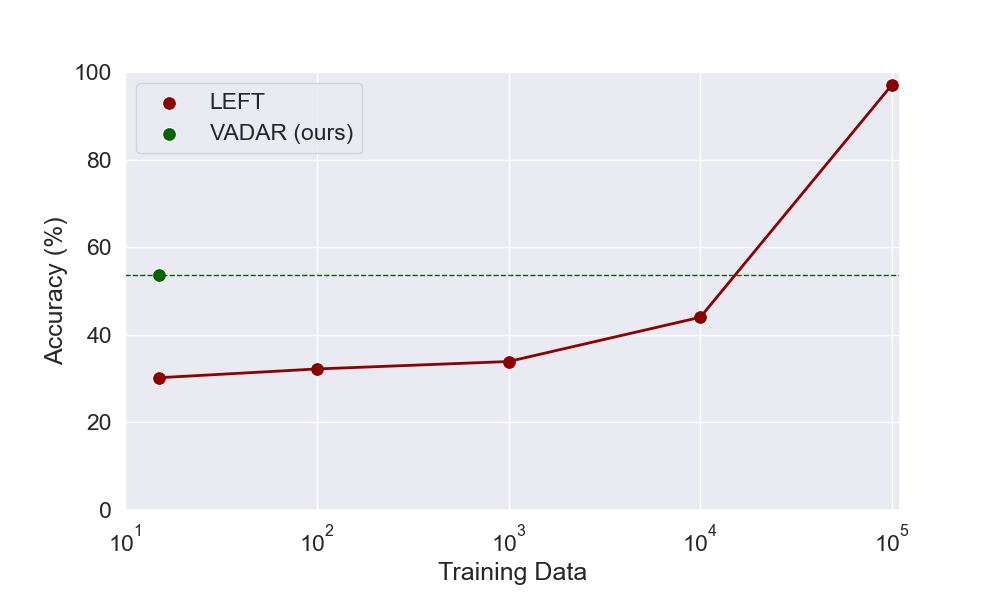}
    \vspace{-8mm}
    \caption{\textbf{\leftm~\cite{whatsleft} vs \method on \clevr.} \leftm requires supervision. We vary the amount of training data (x-axis) and report accuracy (y-axis). \method requires \emph{no} supervision but takes in 15 queries \emph{without answers} to guide the creation of the API. \method outperforms \leftm trained with $\leq 10,000$ supervised examples.}
    \label{fig:left_vs_ours}
    \vspace{-6mm}
\end{figure}

\cref{tab:main_results} compares our approach, \method, to state-of-the-art VLMs and Program Synthesis methods. \cref{fig:left_vs_ours} additionally compares to the neuro-symbolic \leftm~\cite{whatsleft}.
\method uses \gpt4o with a temperature of $0.7$ for all agents.

\mypar{VLMs vs \method.} VLMs, such as \gpt4o~\cite{gpt4}, \claude~\cite{claude}, \gemini~\cite{gemini}, \llama-11B~\cite{llama3}, and Molmo-7B~\cite{molmo}, are monolithic models trained on vast image-question-answer datasets, likely including samples with spatial and 3D information. We expect them to perform well on related tasks. We also compare to SpaceMantis~\cite{spatial, mantis}, the most recent and largest SpatialVLM~\cite{spatial} variant, finetuned on data with 3D information. We analyze performance based on three answer types: yes/no, multiple-choice, and numerical answers. For queries with floating point answers, we report MRA~\cite{thinkinginspace} with thresholds $\mathcal{C} = \{0.5, 0.55,...,0.95\}$ for outputs $\hat{y}$ and ground truth $y$: $\mathcal{MRA} = \frac{1}{|\mathcal{C}|} \sum_{\theta \in \mathcal{C}} \mathbbm{1} \bigg(\frac{|\hat{y} - y|}{y} < 1 - \theta \bigg)$


\begin{figure*}[t!!]
    \centering
    \includegraphics[width=0.49\linewidth]{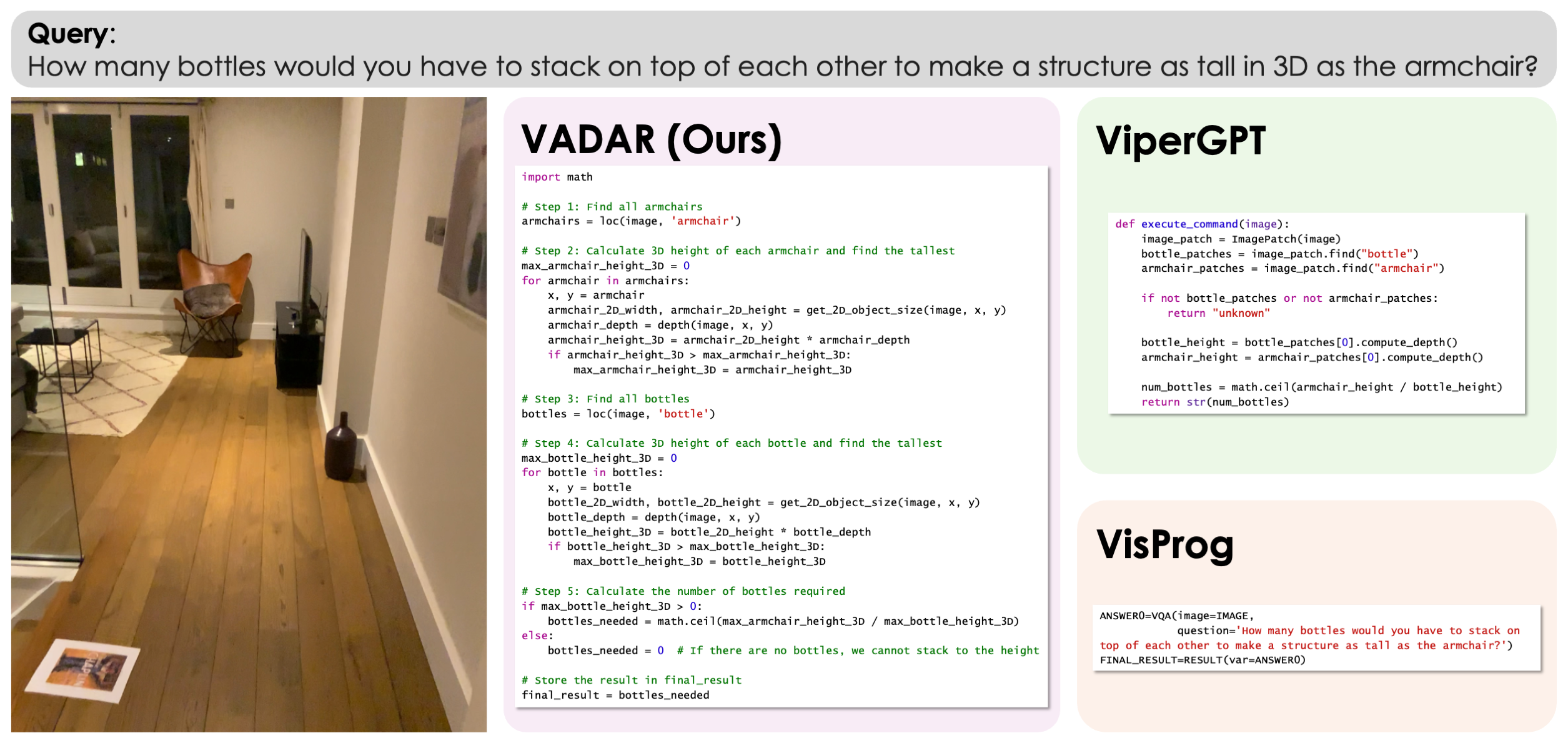}
    \includegraphics[width=0.49\linewidth]{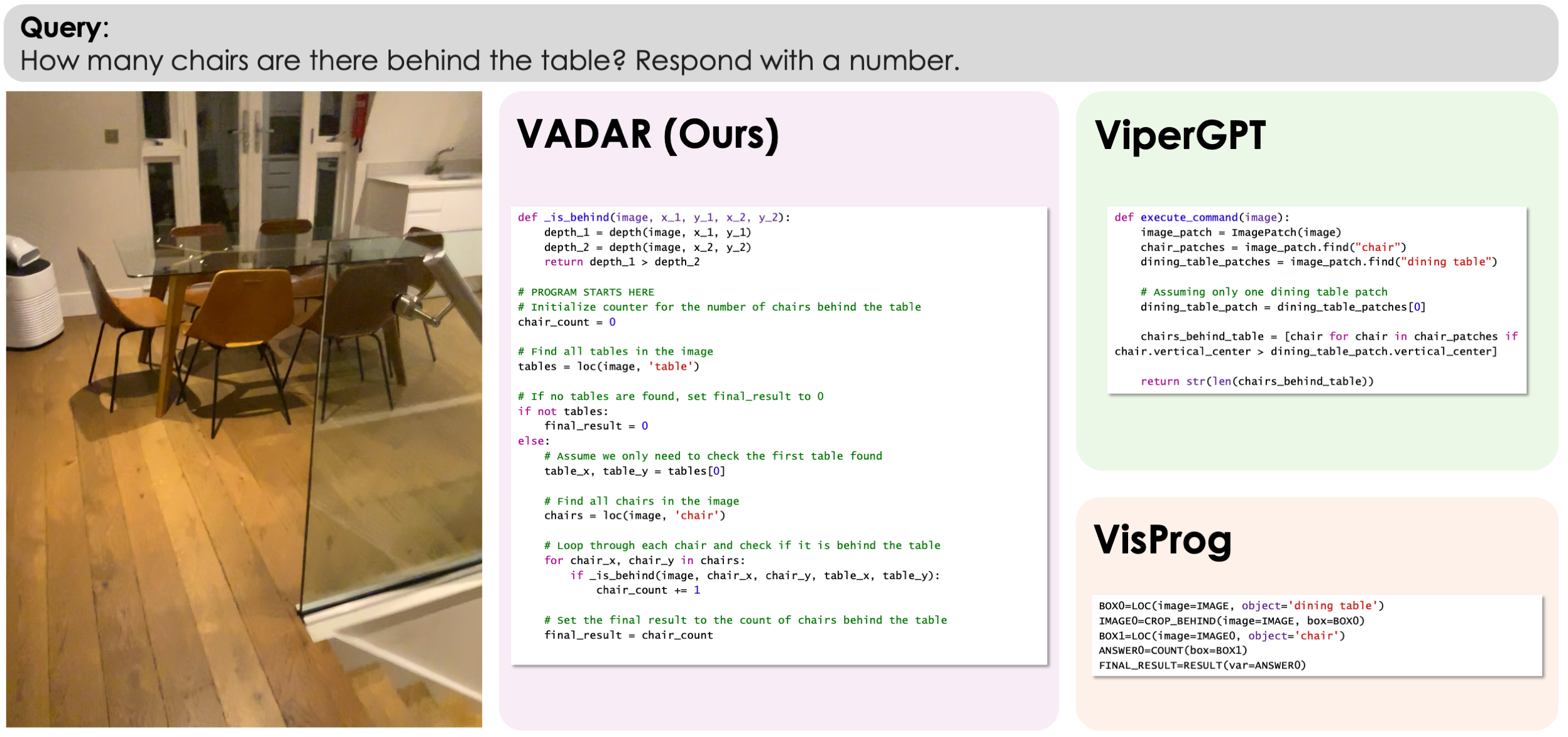}
    \includegraphics[width=0.49\linewidth]{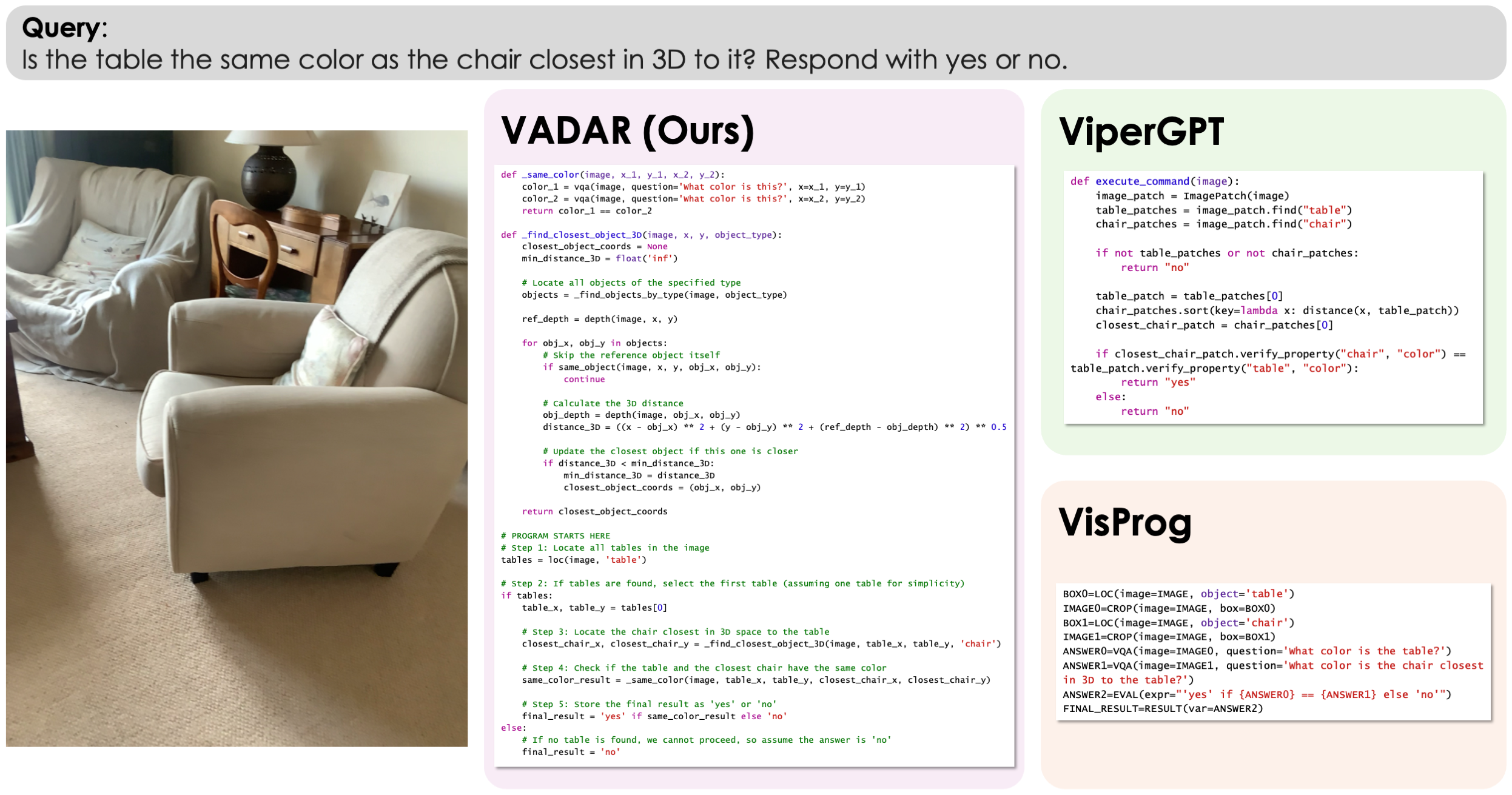}
    \includegraphics[width=0.49\linewidth]{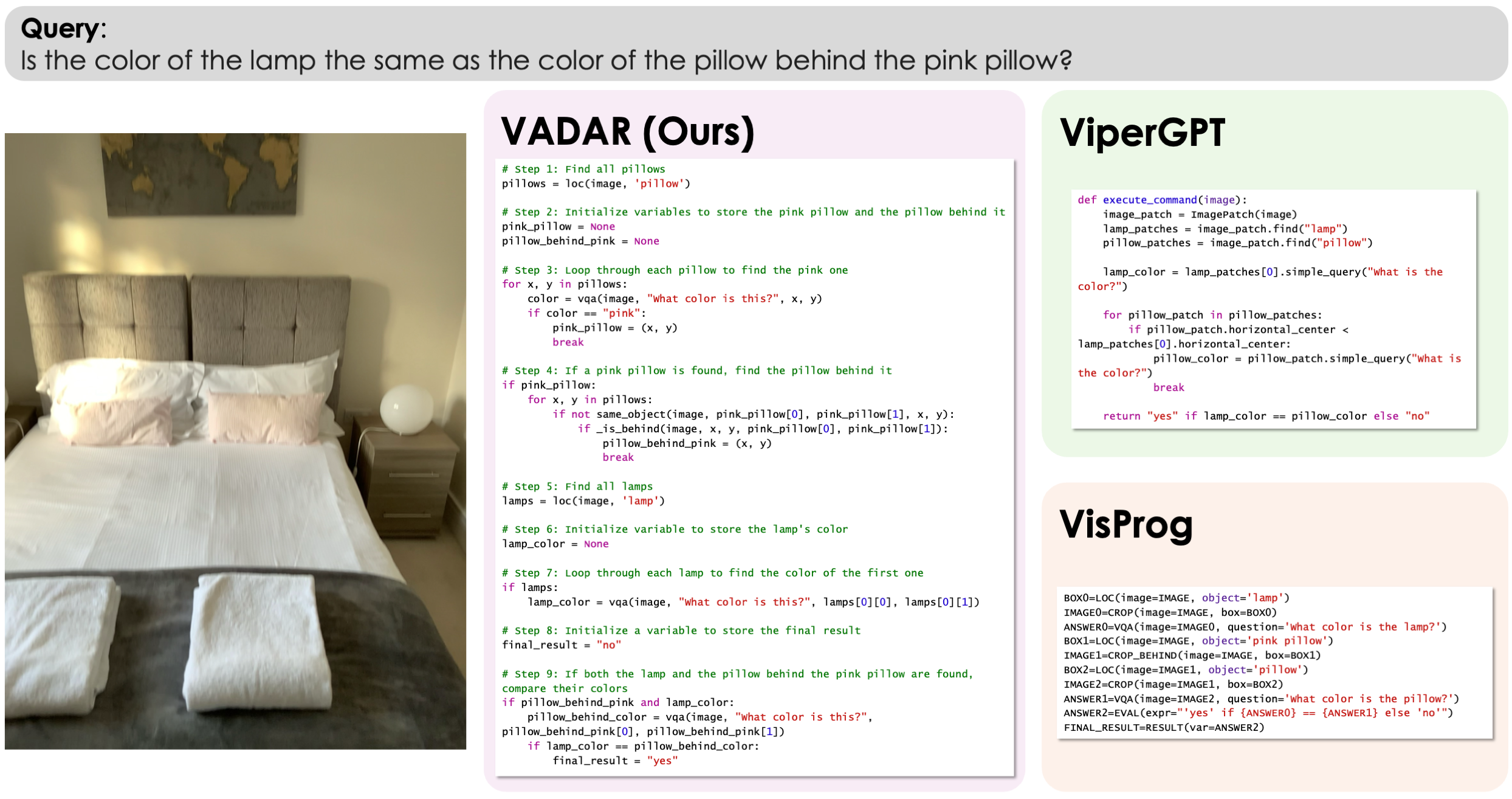}
    \includegraphics[width=0.49\linewidth]{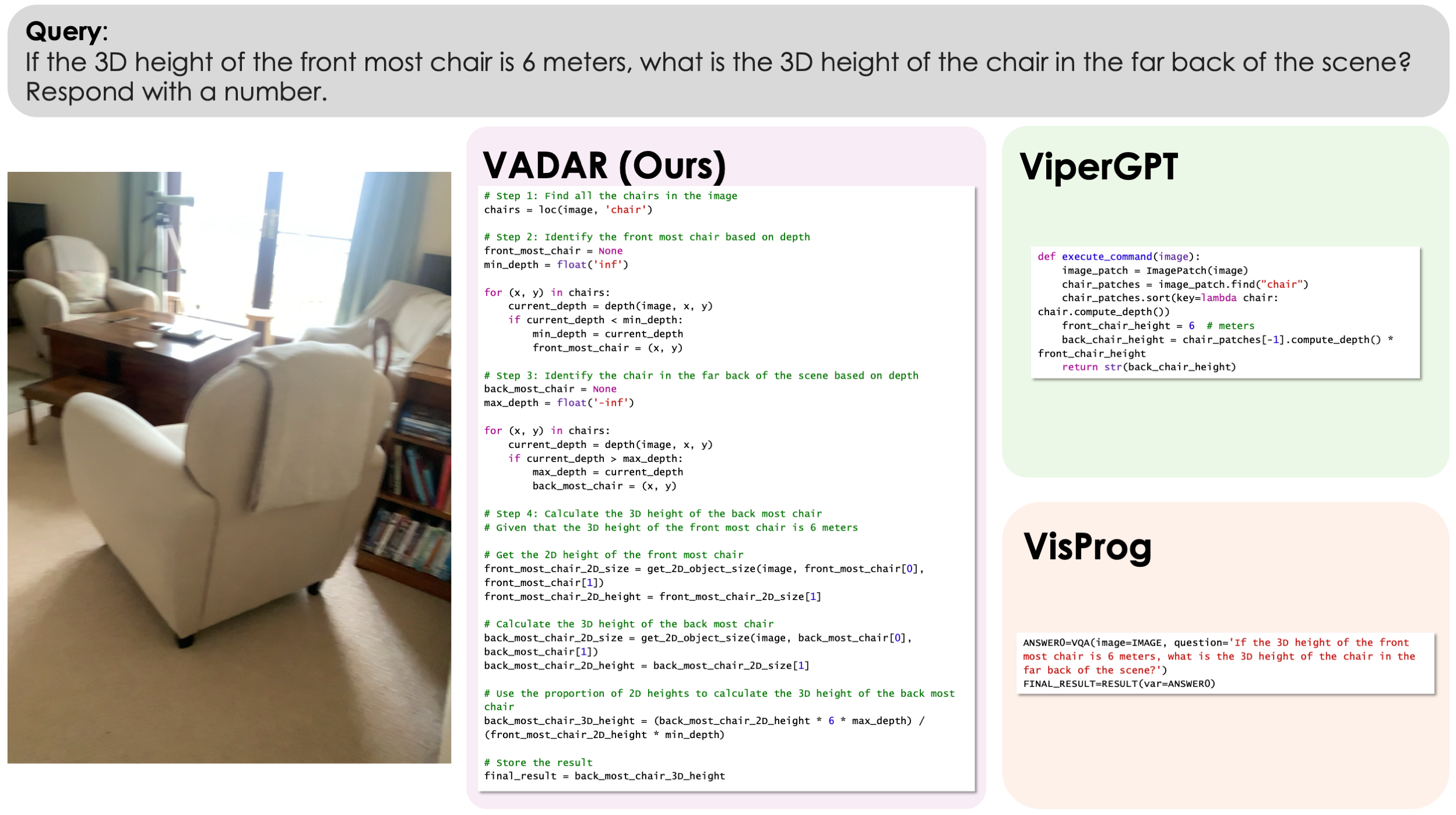}
    \includegraphics[width=0.49\linewidth]{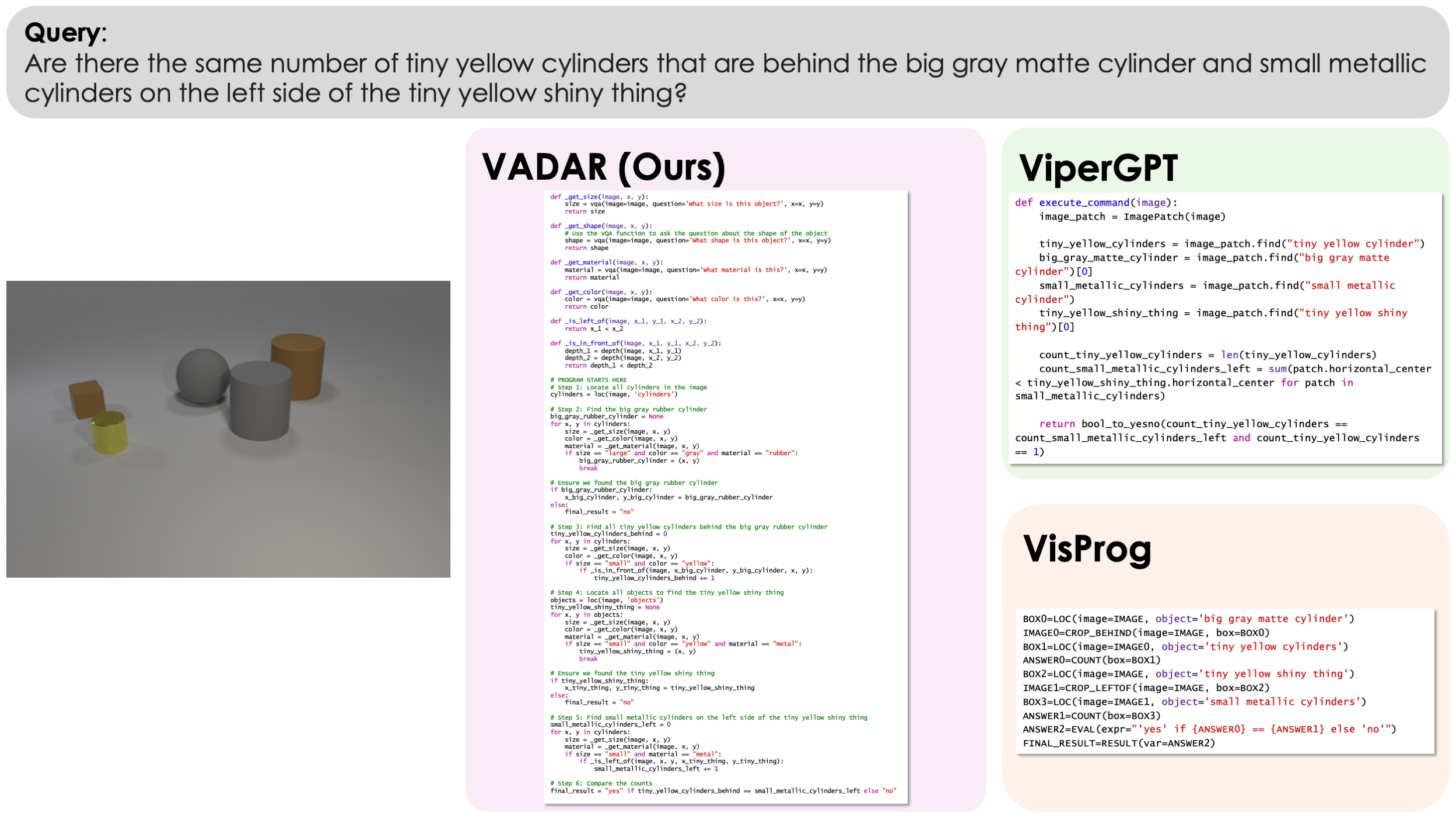}
    \vspace{-3mm}
    \caption{\textbf{Program outputs for \visprog, \viper and \method.} For each example, we show the query, the input image, and the method's program generations. Queries are from our benchmark and pertain to 3D understanding of scenes. Zoom-in to read the programs.}
    \label{fig:main_results}
    \vspace{-5mm}
\end{figure*}

From \cref{tab:main_results}, we observe that
on \clevr, \gpt4o, \claude, and \gemini perform best on average while \method slightly outperforms VLMs on numeric (by 1.0\%) and yes/no answers (by 2.3\%), while providing interpretable execution traces.
On \ourbench, \method is behind GPT4o by just $2\%$ and outperforms all other VLMs by more than $5\%$. \llama-11B and Molmo-7B perform worse among VLMs likely due to their smaller size. 

\mypar{\viper vs \visprog vs \method.} 
\method outperforms both methods on both \clevr and \ourbench by more than 20\%. \visprog and \method use \gpt4o as their LLM; \viper uses GPT-3.5 as it performed better.

Separating program correctness from execution accuracy, \cref{tab:oracle_results} provides comparisons to \viper and \visprog when vision specialists are replaced with oracle ones. 
On \clevr, we use an Oracle Execution Agent that leverages the true scene annotations to provide the correct output automatically.
For \ourbench, we use a smaller subset of 50 queries and manually verify program correctness as ground truth 3D information is not available for all objects in the scene.
The results reveal that with oracle vision specialists, \method achieves an accuracy of 83.0\% on \clevr and 94.4\% on \ourbench, compared to \viper's 42.6\% and 54.9\%, and \visprog's 39.9\% and 66.0\% respectively.
This suggests that \method supports a wider variety of queries, thanks to the dynamically generated API by our LLM agents, as opposed to the static, human-defined API in \viper and \visprog. Our API allows for flexible integration of vision specialists, avoiding human biases -- \eg, as in \visprog, where the pre-defined API guides the LLM to define ``behind'' by cropping the image above.

The high accuracy of \method with oracle vision specialists -- more than 20\% above \claude on \clevr and more than 40\% above \gpt4o on \ourbench -- suggests a promising path to scaling 3D spatial reasoning: improving specialized vision models. These models are easier to train than general-purpose VLMs, as they address simpler tasks with more accessible training data.

\cref{fig:main_results} shows programs generated by the methods. We observe that \viper and \visprog tend to resort to direct VQA calls when questions are complex, as opposed to generating programs. 
In addition, \viper often tends to produce incomplete programs, ignoring a significant portion of the query. 
Finally, both \viper and \visprog often confuse above-behind and below-in front. 
This seems to be a semantic error for \viper that uses a depth estimation module, like us, and a conceptual design error by \visprog that implements \texttt{CROP\_BEHIND} to crop above in the image.

\mypar{\leftm~\cite{whatsleft} vs \method.}
We also compare to the logic-enhanced neuro-symbolic approach \leftm~\cite{whatsleft}, which uses trained modules to ground visual concepts in images, such as ``\texttt{is left of}''. Unlike \leftm, our approach is entirely training-free, while \leftm requires extensive supervision for module training. \cref{fig:left_vs_ours} reports the performance of \leftm on the \clevr dataset when trained (to convergence) with varying training set sizes (x-axis). Although our approach does not require any explicit supervision, our API agent uses a small sample ($=15$) of \emph{questions only}, \emph{without answers}, to construct the API. 
According to \cref{fig:left_vs_ours}, we outperform \leftm trained with $\leq 10,000$ examples on \clevr.
Notably, it is not possible to evaluate \leftm on \ourbench due to its reliance on a large, domain-specific training set with appropriate 3D supervision, which is difficult to obtain for this benchmark or in general. 
This highlights an added advantage of our method: its ability to scale to new domains without the need for training.

\mypar{Results on GQA.}
We report results on GQA~\cite{gqa}, a widely used benchmark for spatial reasoning. 
As noted earlier, GQA queries emphasize object appearance and attributes, and primarily require one-step inference.
Questions in GQA include \emph{“What size is the doughnut the person is eating?”} and \emph{“Who is sitting in front of the water?”}. 
\cref{table:qga} compares \gpt4o, \viper, \visprog, and \method.
We observe different relative model performance compared to \cref{tab:main_results}.
Given the nature of GQA, it is not surprising that a monolithic and performant VLM like \gpt4o would perform well, which our results confirm. Among the program synthesis methods, we observe that \method and \visprog achieve comparable performance, while \viper shows a drop in accuracy. 
A deeper dive into the output programs shows that \visprog relies on image-wide VQA calls in 34\% of cases, whereas \method does so only 24\% of the time.
The limitations of GQA queries in evaluating 3D spatial reasoning highlight the need for our proposed benchmark, which better assesses 3D understanding and exposes the weaknesses of current methods.
\begin{table}[]
\begin{center}
\begin{tabular}{l|c}
Method  & GQA \\
\hline
\gpt4o~\cite{gpt4}      & \bf{54.9} \\
\viper~\cite{vipergpt}  & 42.0  \\
\visprog~\cite{visprog} & 46.9 \\
\method (ours)          & 46.1 
\end{tabular}
\end{center}
\vspace{-5mm}
\caption{\textbf{Results on GQA} on a subset of testdev split. GQA focuses primarily on object appearance, not 3D spatial reasoning.}
\label{table:qga}
\vspace{-4mm}
\end{table}

\subsection{Ablations}
We turn to ablations to quantify the effectiveness of the agentic design and prompting in our approach. To reduce costs from \gpt4o, we experiment on a randomly selected CLEVR subset. \cref{table:ablations_agents} compares the following variants:

\begin{table}[]
\begin{center}
\begin{tabular}{l|c}
\multicolumn{1}{c|}{}  & \clevr$_{100}$  \\
\hline
No-API Agent    & 60.7 \\
API Agent       & 64.0  \\
+ Weak ICL      & 65.7  \\
+ Pseudo ICL    & 66.7  
\end{tabular}
\end{center}
\vspace{-5mm}
\caption{\textbf{Ablations of agentic design and prompts} on \clevr$_{100}$, a subset of 100 questions. We compare to single agent variant \emph{No-API} which creates programs directly. We then ablate prompting by incrementally adding instructions to the agents used to define the API. The No-API Agent performs the worst and our prompting techniques add to \method's performance.}
\label{table:ablations_agents}
\vspace{-4mm}
\end{table}

\begin{figure}
    \centering
    \includegraphics[width=\linewidth]{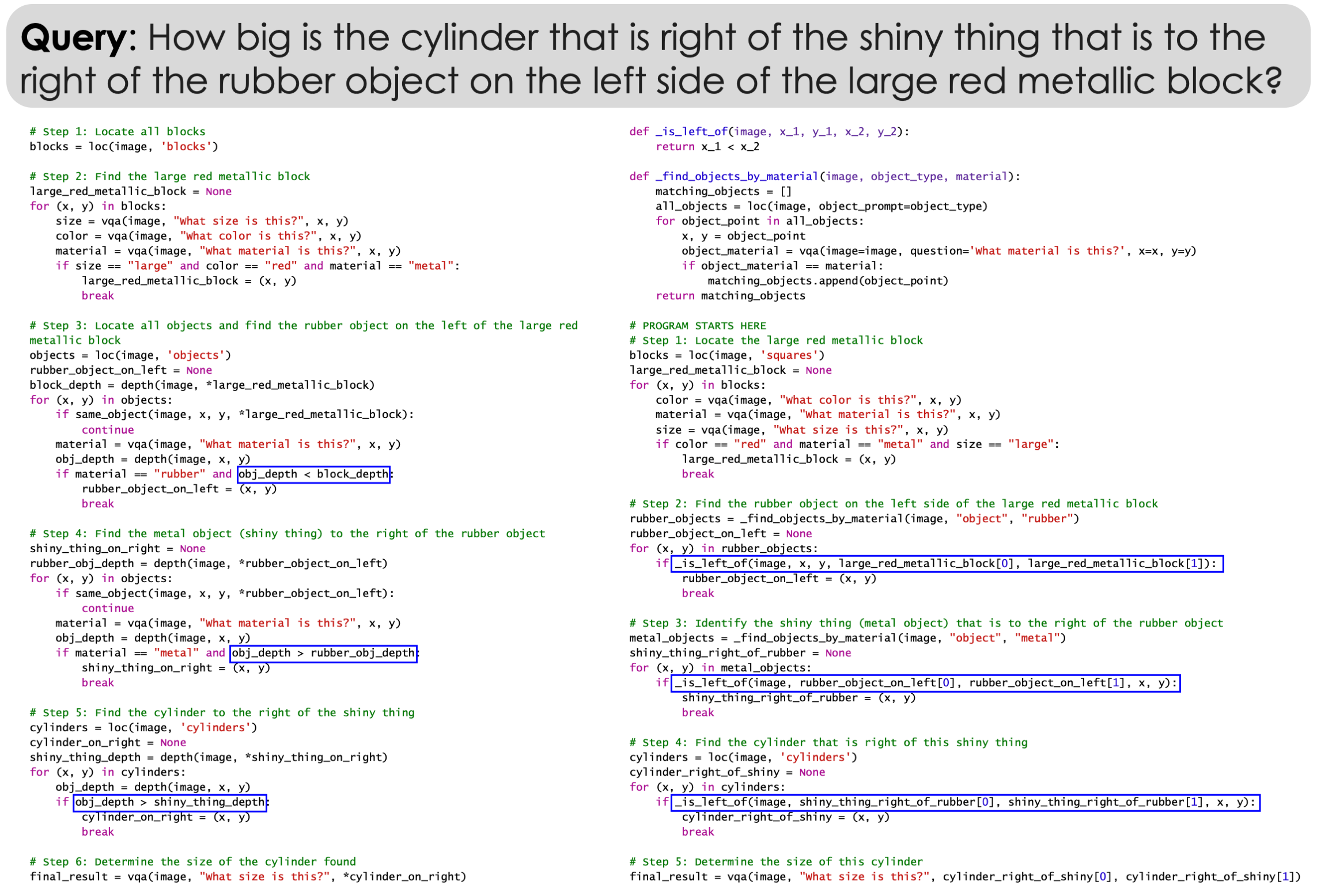}
    \begin{minipage}{0.4\linewidth}
    \centering
    (a) \emph{No-API} Agent
    \end{minipage}
    \begin{minipage}{0.58\linewidth}
    \centering
    (b) \method
    \end{minipage}
    \vspace{-3mm}
    \caption{(a) The \emph{No-API} agent produces longer programs and is prone to errors, often mistakenly using depth for left/right comparisons. (b) In contrast, our agentic \method creates shorter programs by leveraging methods from the API.}
    \label{fig:noapi_vs_ours}
    \vspace{-6mm}
\end{figure}

\myparit{No-API Agent} is a single agent instructed to directly create programs for queries without defining an API of reusable methods. Comparison to this variant shows the value of an API. 
\cref{fig:noapi_vs_ours} shows a common reasoning error by the \emph{No-API Agent}, which confuses depth with left/right; our approach, by implementing reusable methods, invokes the appropriately named method that is accurately implemented.
The example reiterates that spatial reasoning relies on correctness, supporting \method's design to build an accurate API \textit{before} program synthesis, over library learning, that discovers a potentially incorrect library \textit{after} program synthesis.

\myparit{API Agent} is our approach without any prompting instructions or ICL examples.
We incrementally add our two prompting techniques: 
(1) \emph{Weak ICL} examples guide the Implementation Agent to use the pre-defined modules. 
(2) \emph{Pseudo ICL} provides pseudo-code examples and instructions in \emph{natural language} to the Implementation and Program Agent, respectively, that demonstrate how to handle intricate queries. We provide the prompts in the Appendix. 

From \cref{table:ablations_agents} we observe that the No-API Agent performs the worst, while our prompting techniques via weak ICL examples and instructions achieve the best performance.

\section{Limitations \& Future Work}
\label{sec:conclusion}
We introduce \method, an agentic approach that leverages LLM agents to dynamically create and expand a Pythonic API for complex 3D visual reasoning tasks. Our agents autonomously generate and implement functions, which are then utilized by the Program Agent to produce programs. This reuse of functions results in more accurate programs for complex queries.
There is an extensive list of future directions to address current limitations of \method.
\begin{itemize}
\item \method often struggles with queries that require 5 or more inference steps, \eg \emph{“There is a yellow cylinder to the right of the cube that is behind the purple block; is there a brown object in front of it?”}. We provide the programs for these complex cases in the Appendix. Addressing such queries can be improved by leveraging advanced prompting strategies, an active research area that includes methods like CoT~\cite{cot} and prompt chaining~\cite{promptchainer, aichains}.
\item We show that \method attains high program accuracy (\eg, 83.0\% on \clevr) but lower execution accuracy (53.6\%) due to errors from the vision specialists. A potential enhancement would be to enable \method to dynamically choose its vision modules from a pool of available options based on empirical performance. Integrating the selection process with reinforcement learning or self-improvement mechanisms is a promising future direction.
\item \method creates a program based solely on the input query, utilizing the image only during execution. Incorporating the image into the program synthesis process could improve accuracy, potentially improving performance on queries requiring five or more inference steps.
\end{itemize}
\section*{Acknowledgments}
The project is funded by Meta through the LLM evaluation research grant and partly through Caltech's CAST program.
We also thank Google's Gemma Academic program and OpenAI for granting us API credits for their LLMs.

{
    \small
    \bibliographystyle{ieeenat_fullname}
    \bibliography{main}
}

\clearpage
\setcounter{page}{1}
\maketitlesupplementary

\appendix
\begin{table}[t!]
\centering
\resizebox{1.0\columnwidth}{!}{
\begin{tabular}{ll|c|c}
                                & \multicolumn{1}{c|}{Method}              & \multicolumn{1}{c|}{\clevr}         & \multicolumn{1}{c}{\ourbench}             \\
\hline
\multirow{6}{*}{\rotatebox[origin=c]{90}{\small VLMs}} & GPT4o~\cite{gpt4}  & 1.4 & 0.6 \\
                                & Claude3.5-Sonnet~\cite{claude}      & 0.2 & 0.6 \\
                                & Llama3.2~\cite{llama3}              & 0.5 & 1.6 \\
                                & Gemini1.5-Pro~\cite{gemini}         & 0.3 & 1.8\\
                                & Gemini1.5-Flash~\cite{gemini}       & 0.3 & 1.1 \\
                                & Molmo~\cite{molmo}                  & 0.0 & 0.0 \\
                                & SpaceMantis~\cite{mantis, spatial}  & 0.0 & 0.0 \\
\hline
\multirow{3}{*}{\rotatebox[origin=c]{90}{\small \shortstack{Program\\Synthesis}}} 
                                & ViperGPT~\cite{vipergpt}            & 1.1 & 0.3 \\
                                
                                & VisProg~\cite{visprog}              & 0.9 & 0.3 \\
                                & \method (ours)                      & 2.9 & 1.8 \\
\end{tabular}}
\vspace{-2mm}
\caption{\textbf{Standard deviation across experimental runs.} \method's variation is comparable to VLMs on Omni3D, but slightly higher than program synthesis methods on \clevr, despite achieving significantly higher accuracy.}
\label{tab:stdevs}
\vspace{-3mm}
\end{table}

\begin{table}[h!]
\centering
\resizebox{0.99\columnwidth}{!}{
\begin{tabular}{cccc}
Signature (for 10 Qs) & Implementation & Program (per Q) & Execution (per Q)\\ \hline
$20.5_{\pm 3.6}$ & $37.2_{\pm 14.4}$ & $6.5_{\pm 1.8}$ & $35.7_{\pm 11.8}$ 
\end{tabular}
}
\caption{\textbf{Runtime for each Agent in seconds.}}
\label{tab:runtime}
\end{table}

The Appendix includes the prompts used for all agents, additional qualitative examples of \method on \clevr, \ourbench, and GQA, and a supplemental qualitative analysis with standard deviations to compare the robustness of approaches.

\section{Prompts}
\mypar{Predefined Module Signatures.} \cref{fig:clevr_predefined} and \cref{fig:omni3d_predef} show the docstrings of the predefined modules 
for \clevr and \ourbench respectively, which are used to initialize the dynamic API. We note that the two prompts are virtually identical, with the exception of the \texttt{get\_2D\_object\_size} method, which we omit from our experiments on \clevr as the dataset defines size as either \texttt{small} or \texttt{large}. In~\cref{fig:predefined_implementation}, we provide the Python implementation for all of the predefined modules.

\mypar{Signature Agent Prompt.} \cref{fig:signature_agent_prompt} contains the prompt used for the Signature Agent for both \clevr and \ourbench. We prompt the LLM to only generate signatures for methods when necessary, as we found this avoids redundant methods with minor changes to previously defined methods. We impose that the name of new methods start with an underscore, to prevent the common failure case of methods sharing names with variables previously defined. 

\mypar{Implementation Agent Prompt.} \cref{fig:implementation_prompt_clevr} and \cref{fig:implementation_prompt_omni3d} contain the prompts used for the Implementation agent on \clevr and \ourbench respectively. The prompts contain \emph{Weak ICL} examples, illustrating how to implement a model signature and use the pre-defined modules correctly for simpler queries. This is in contrast to \emph{Strong ICL} examples in \visprog and \viper, which provide complete program examples for full queries using a predefined API. In our framework, where agents dynamically generate the API, \emph{Strong ICL} is not feasible. 

Additionally, the prompts feature \emph{Pseudo ICL} in the form of natural language instructions and tips. Similarly to the predefined modules, the prompts differ between \clevr and \ourbench as the latter considers metric sizes and not a binary \texttt{small} or \texttt{large} as in \clevr. Consequently, we found it necessary to include natural language definitions and instructions for reasoning about 2D and 3D dimensions in the Implementation prompt on \ourbench.

\mypar{Program Agent Prompt.} In \cref{fig:program_prompt_clevr} and \cref{fig:program_prompt_omni3d} we show the prompts for the Program Agent on \clevr and \ourbench respectively. In the prompt for \clevr, we include a list of all available attributes. In both prompts, we include \emph{Pseudo ICL} in the form of natural language examples and instructions. For the \ourbench prompt, we additionally include tips and definitions for handling 2D and 3D dimensions.

\section{Additional Quantitative Analysis}
\mypar{Experimental Variability.} \cref{tab:main_results} in the main paper reports the mean performance of all methods across $3$ runs. \cref{tab:stdevs} reports the standard deviation on \clevr and \ourbench across the same $3$ runs. \method's variation is comparable to the VLMs on \ourbench, but slightly higher than program synthesis methods on both benchmarks. However, \method significantly outperforms ViperGPT and VisProg, even when accounting for this variation.

\mypar{Runtime.} \cref{tab:runtime} reports runtime in seconds for our Agents on an A100 GPU. Notably, when running our method on $1000$+ questions, the Signature and Implementation Agents \emph{only run once}, therefore their runtime becomes negligible to total inference runtime.

\section{More information on \ourbench}
On images sourced from Omni3D~\cite{omni3d} we collect a set of challenging questions with the help of human annotators. We omit using templates for questions, as done by others~\cite{cambrian1,thinkinginspace,spatial}, to avoid template overfitting, and instead instruct annotators to directly ask questions in free-form natural language, focusing on the scene, object layout and object sizes. 
We discard questions that are simplistic, \eg ``Is there a sofa in the image?" or ``Is the sofa behind the table?", and only keep queries which involve complex inference steps in 2D and 3D.
\ourbench queries roughly target the following areas of reasoning: relative size and dimensions with hypotheticals, spatial relationships and depth reasoning, relative proportions and alignments, and interaction with other objects. Queries from \ourbench can be browsed in \href{https://glab-caltech.github.io/vadar/omni3d-bench.html}{https://glab-caltech.github.io/vadar/omni3d-bench.html}.

We compute answers for questions using the 3D annotations provided in Omni3D~\cite{omni3d}.
Since the questions are not templated and thus don't follow rule-based instructions, we collect answers manually by sourcing the 3D annotations provided by the dataset for each image.
This results in 500 \emph{unique} and challenging image-question-answer tuples that test diverse aspects of 3D spatial reasoning. 
The diversity and complexity of \ourbench is showcased by the examples in~\cref{fig:fig1}, \cref{fig:main_results} and \cref{fig:omni_examples}.

\ourbench complements \clevr when assessing 3D spatial understanding. While \clevr uses templated questions, enabling the creation of a large volume of image-question-answer pairs, \ourbench focuses on diverse and complex reasoning tasks in free-form language. 
Together, \clevr and \ourbench provide a comprehensive test for models' 3D spatial reasoning capabilities. This is evidenced by the relatively low performance of modern state-of-the-art AI models on these benchmarks, achieving only 20-40\% accuracy.

\section{Comparison to VSI-Bench}
\label{sec:vsi-bench}
\begin{table}[t!]
\centering
\resizebox{0.6\columnwidth}{!}{
\begin{tabular}{c|c}
              & VSI-Bench-img \\ \hline
Gemini1.5-Pro & 49.5            \\
\method         & \textbf{50.1}  
\end{tabular}}
\vspace{-2mm}
\caption{\textbf{Results on VSI-Bench~\cite{thinkinginspace}.} \method outperforms Gemini1.5-Pro on a image-based subset of 75 queries from VSI-Bench that sources the frame that contains all the information necessary to respond correctly. Notably, \method achieves a 50.1\% accuracy on this subset, compared to 40.4\% on \ourbench, highlighting the challenging nature of our proposed benchmark.}
\vspace{-3mm}
\label{tab:vsi-bench-results}
\end{table}

Concurrent to our work is VSI-Bench~\cite{thinkinginspace}, a video understanding benchmark that focuses on spatial reasoning. 
VSI-Bench targets 3D reasoning, but it differs from \ourbench in three critical ways:
First, it focuses on video understanding and retrieving the appropriate frame to answer a given query.
Second, while queries in VSI-Bench target 3D object attributes, they query absolute measurements, such as \emph{``What is the height of the chair?"}.
Monolithic VLMs when prompted with such questions resort to object priors. For example, \gpt4o says: \emph{``A chair tends to be 30-40 inches tall"}. 
In contrast, \ourbench introduces hypotheticals that require reasoning over scene attributes, evaluating true 3D spatial reasoning, \eg, \emph{``If the table is 2 meters wide, how tall is the chair?"}.
Third, VSI-Bench queries are templated, which can lead to biased conclusions due to template overfitting.

We compare \method on VSI-Bench. To decouple frame retrieval from image-based reasoning, we create a variant of the benchmark by sourcing a subset of $75$ queries with the associated frame that contains the information necessary to address the query.
We call this subset VSI-Bench-img. 
\cref{tab:vsi-bench-results} reports \method's performance and compares to Gemini1.5-Pro, which authors report to be the best VLM on the set.
From \cref{tab:vsi-bench-results} we observe that \method performs on par with the industry-leading Gemini1.5-pro. 
Importantly, \method's performance on VSI-Bench-img is 10\% higher than on \ourbench (40.4 vs 50.1) which highlights the more challenging nature of our benchmark.

\section{Qualitative Examples on \clevr}

\begin{figure*}[!ht]
    \centering
    \includegraphics[width=0.49\linewidth]{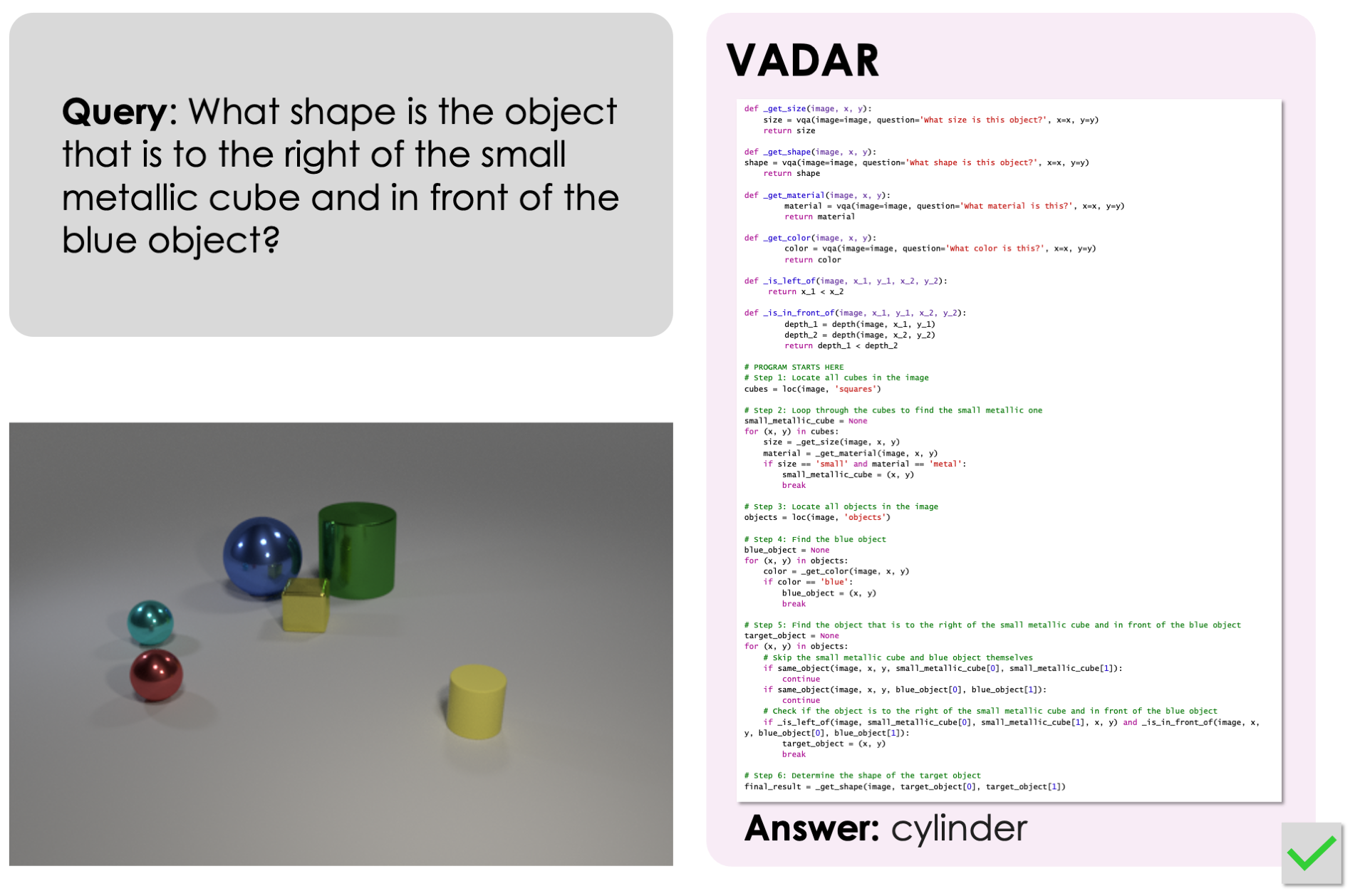}
    \includegraphics[width=0.49\linewidth]{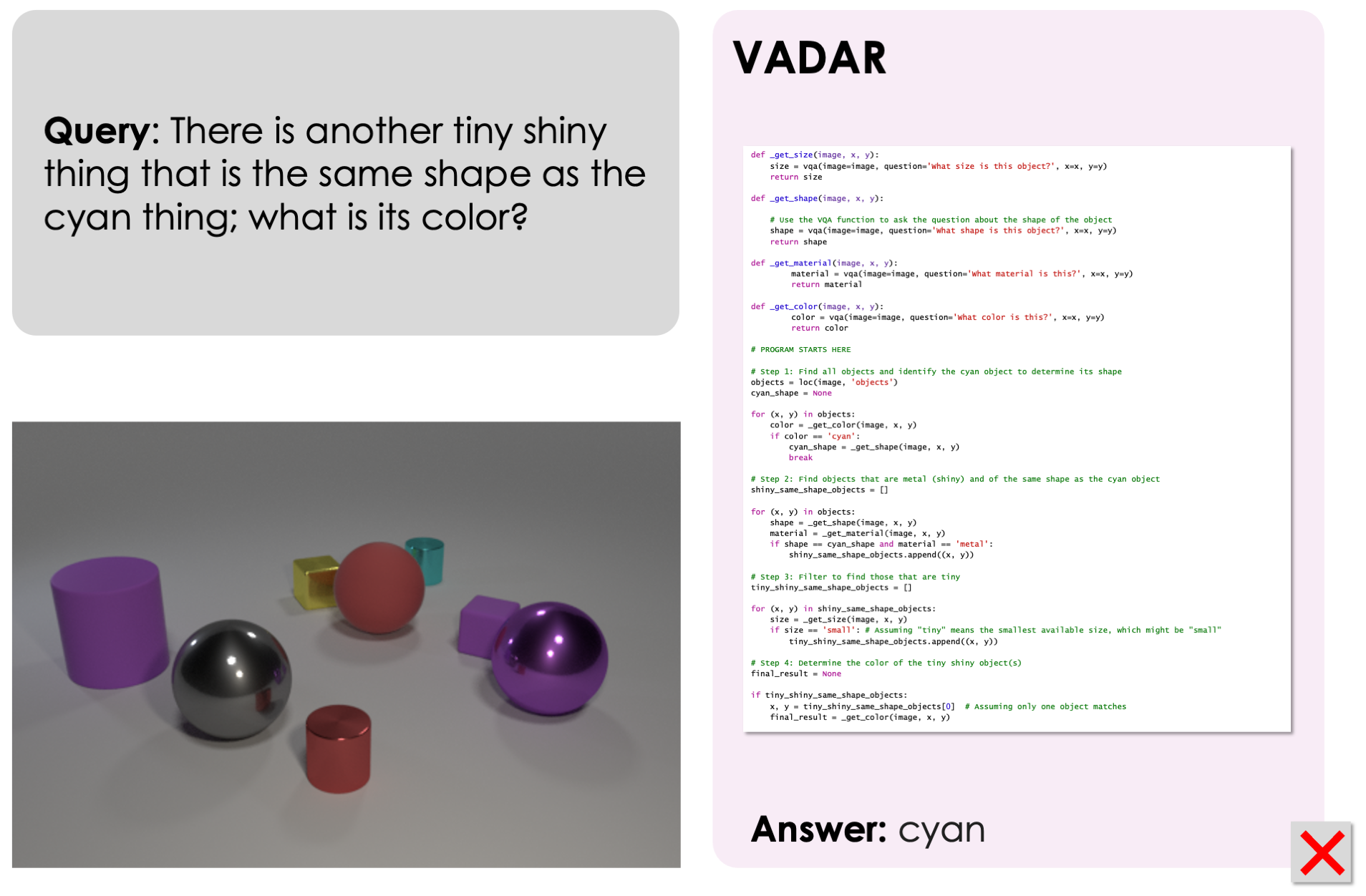}
    \vspace{-2mm}
    \caption{\method program outputs on \clevr.}
    \label{fig:clevr_examples}
    \vspace{-2mm}
\end{figure*}

\cref{fig:clevr_examples} shows additional qualitative examples on \clevr. The correct example showcases the use of API methods for repeated tasks and accurately determining spatial relations. The incorrect example highlights a failure to use same object to exclude the original reference object when the questions asks for ``another'' object.

\section{Qualitative Examples on \ourbench}

\begin{figure*}[!ht]
    \centering
    \includegraphics[width=0.32\linewidth]{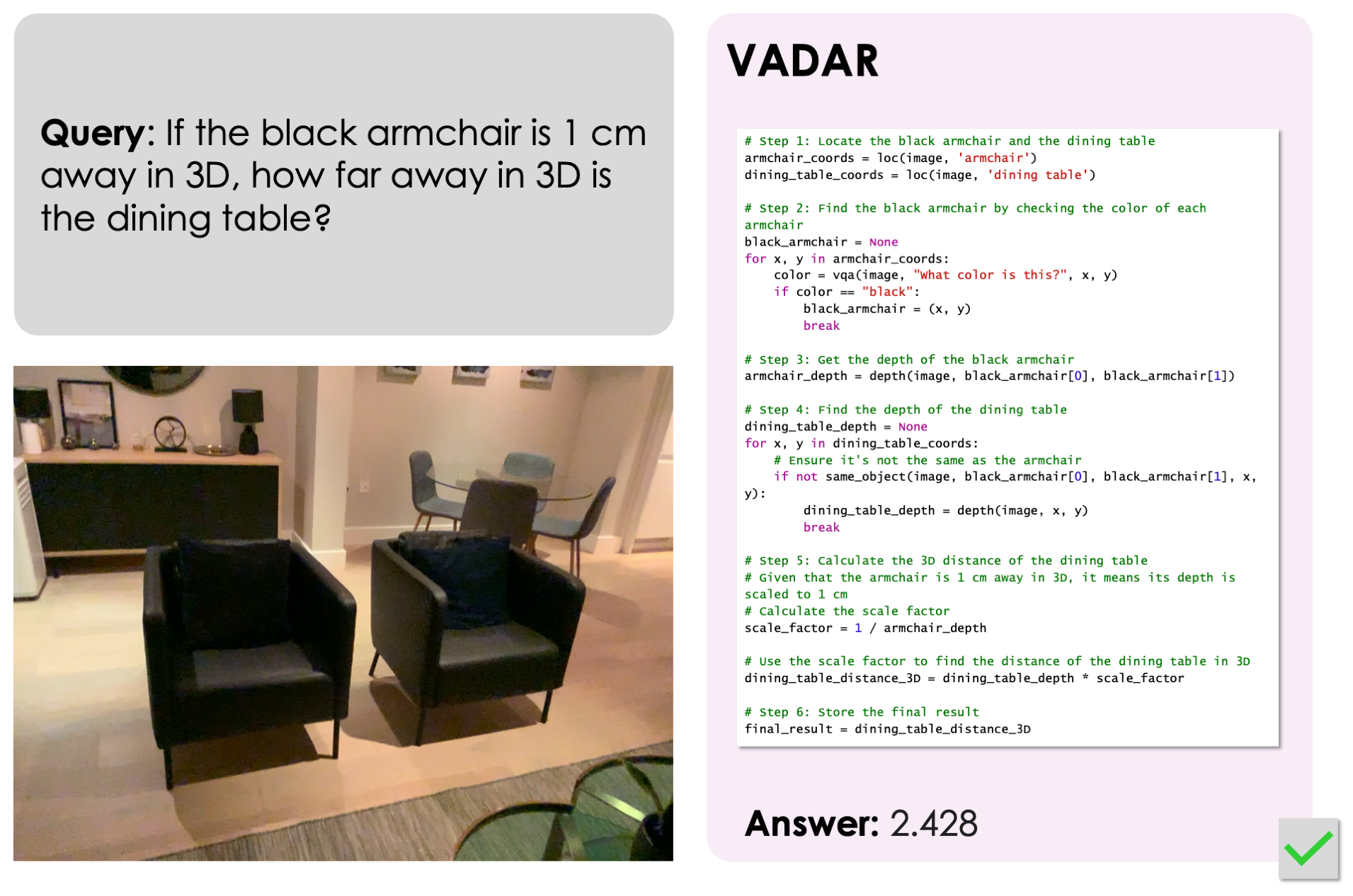}
    \includegraphics[width=0.32\linewidth]{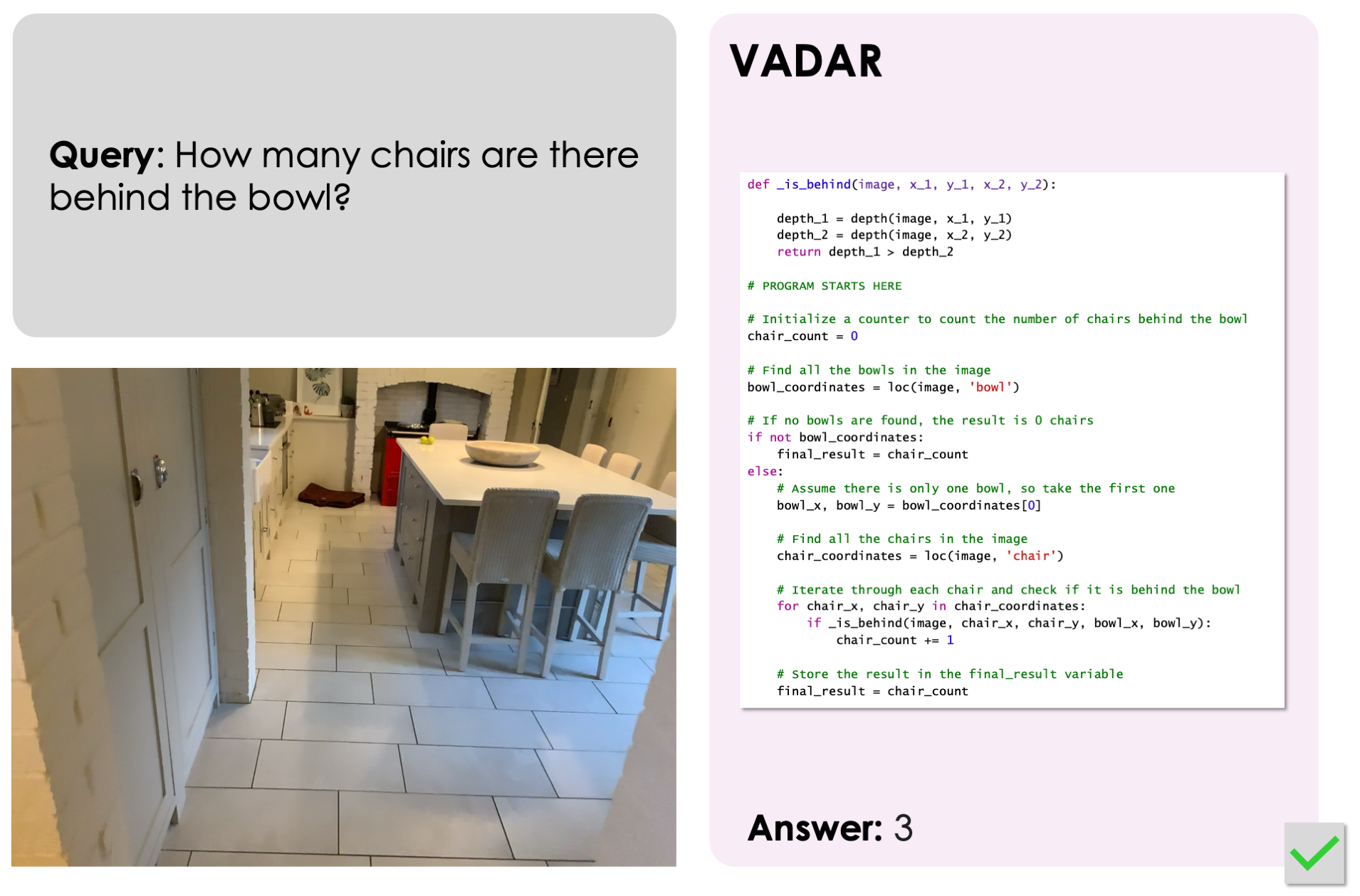}
    \includegraphics[width=0.32\linewidth]{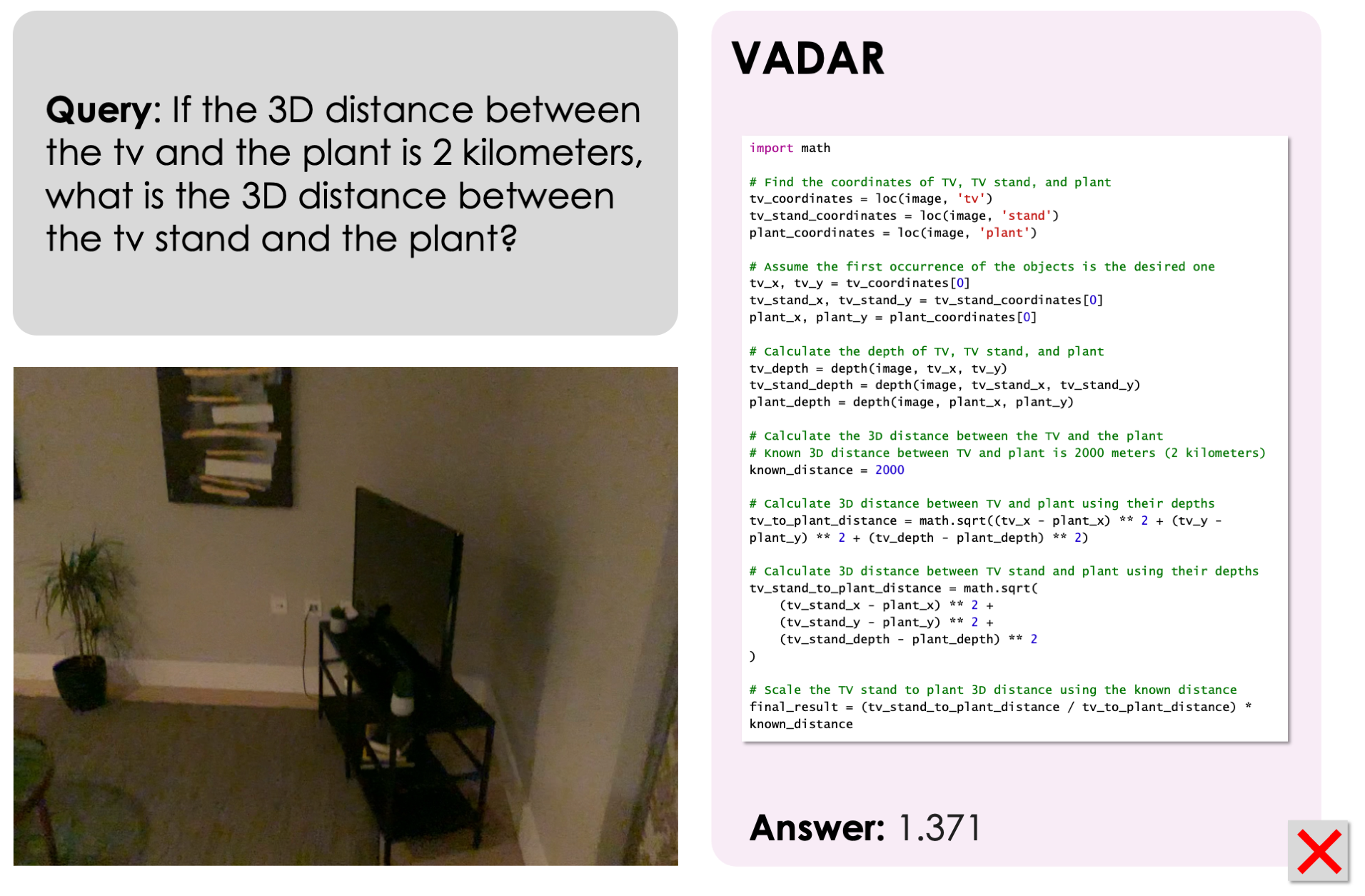}
    \vspace{-2mm}
    \caption{\method program outputs on \ourbench.}
    \label{fig:omni_examples}
    \vspace{-2mm}
\end{figure*}

\cref{fig:omni_examples} shows additional qualitative examples on \ourbench. Our method is able to correctly estimate 3D distances by scaling depth based on the reference scale given in the question. An instance where such scaling is done incorrectly is shown in the last example. 

\section{Qualitative Examples on GQA}

\begin{figure*}[!ht]
    \centering
    \includegraphics[width=0.49\linewidth]{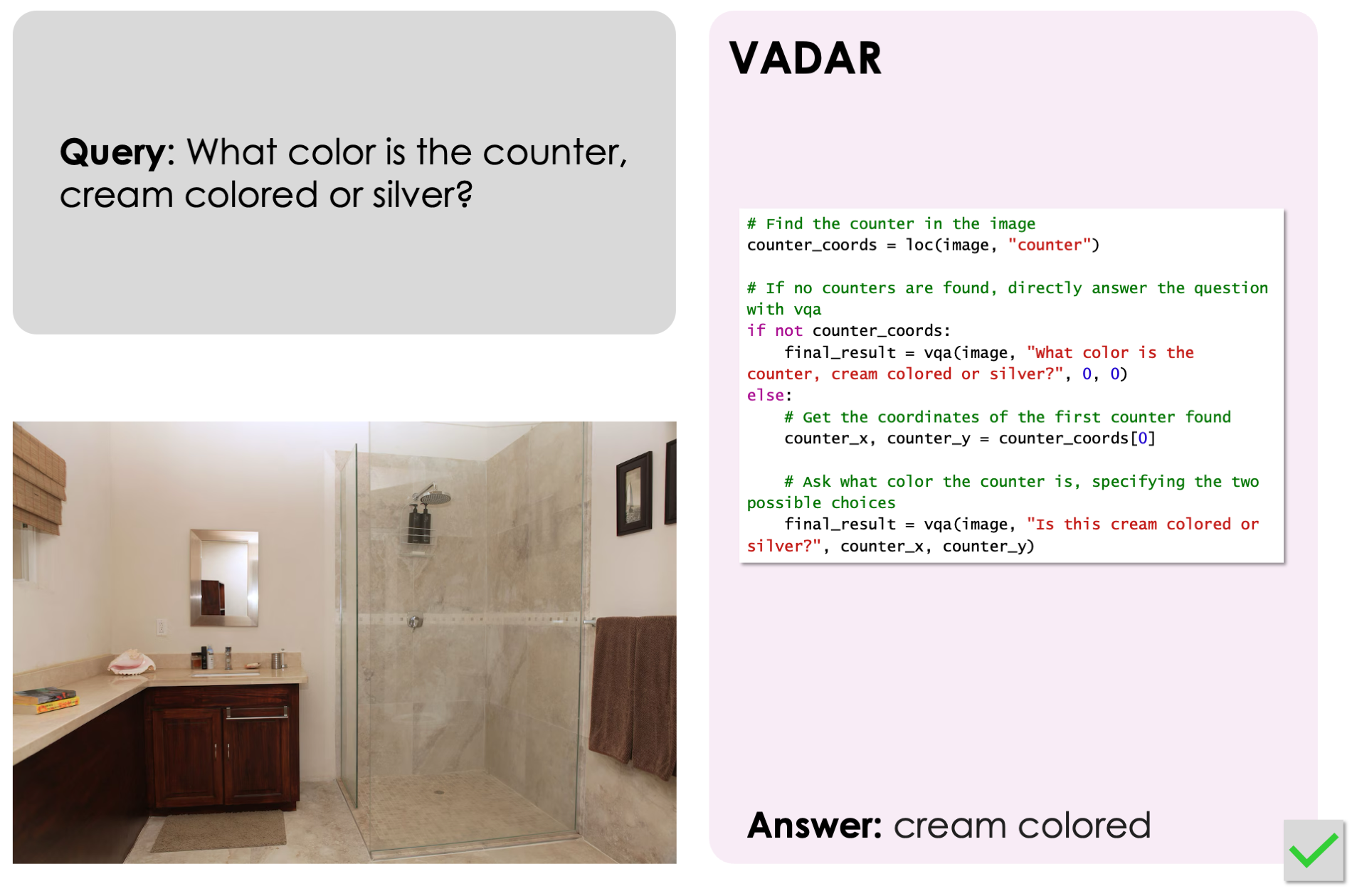}
    \includegraphics[width=0.49\linewidth]{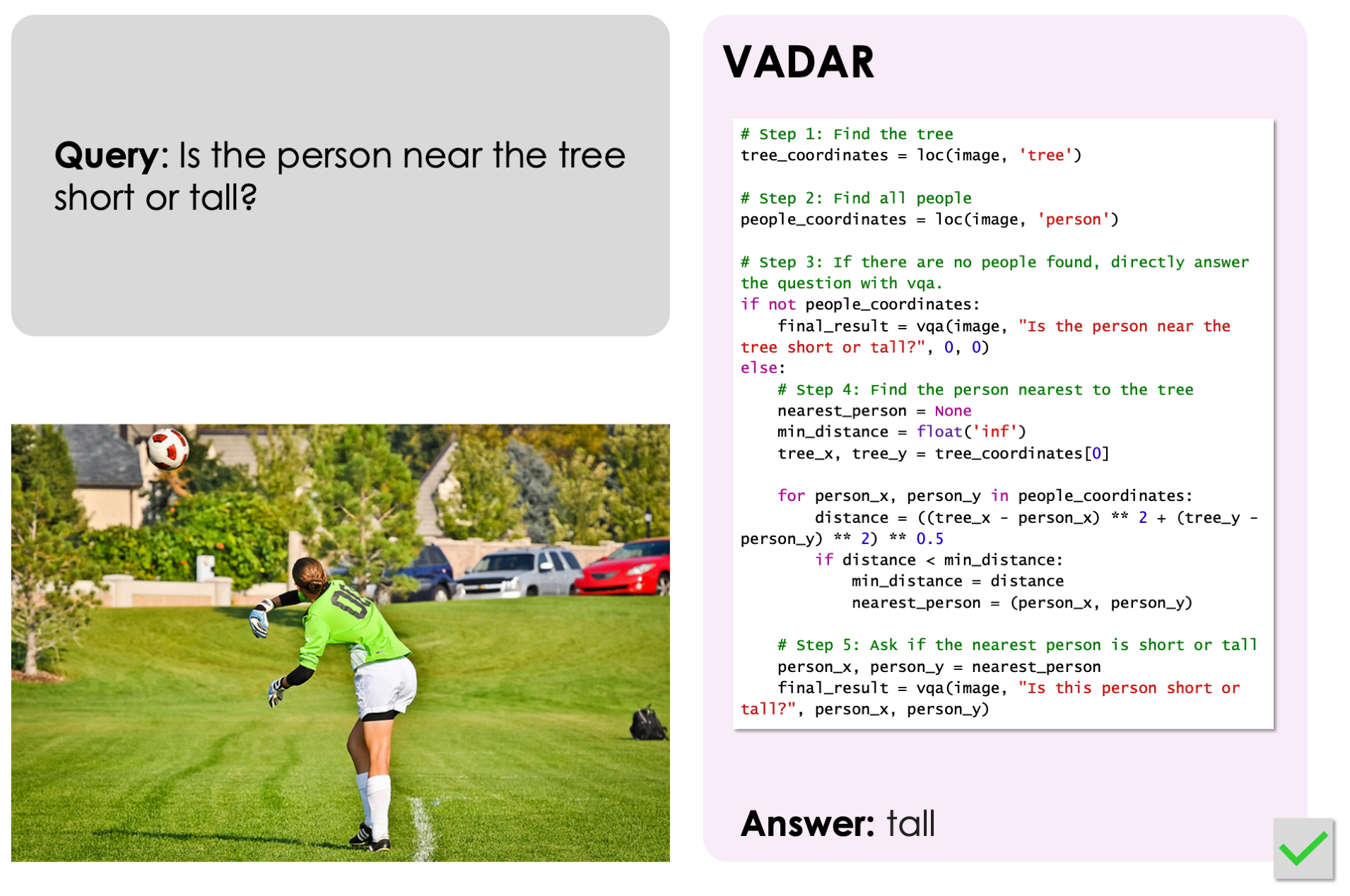}
    \includegraphics[width=0.49\linewidth]{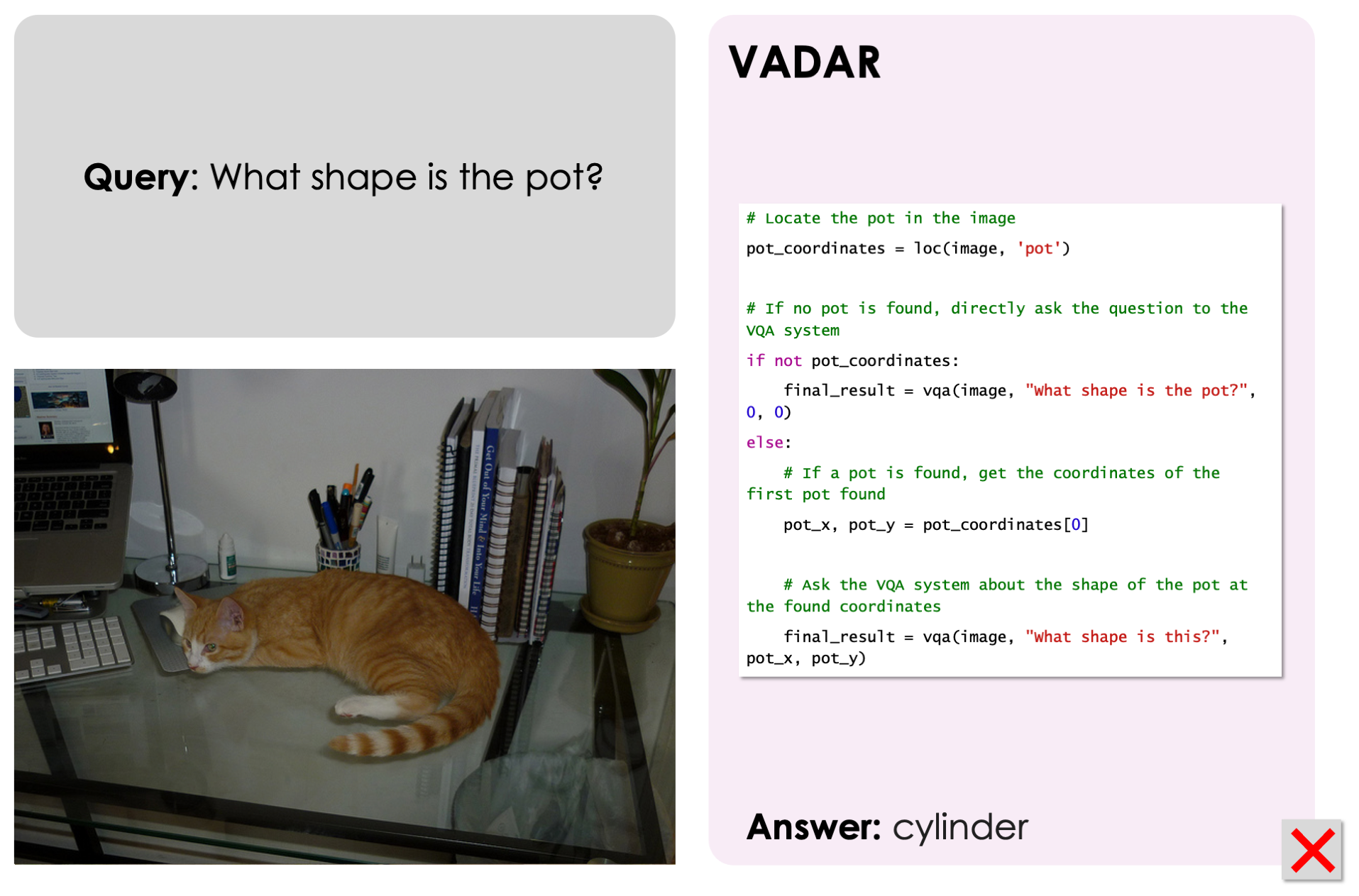}
    \includegraphics[width=0.49\linewidth]{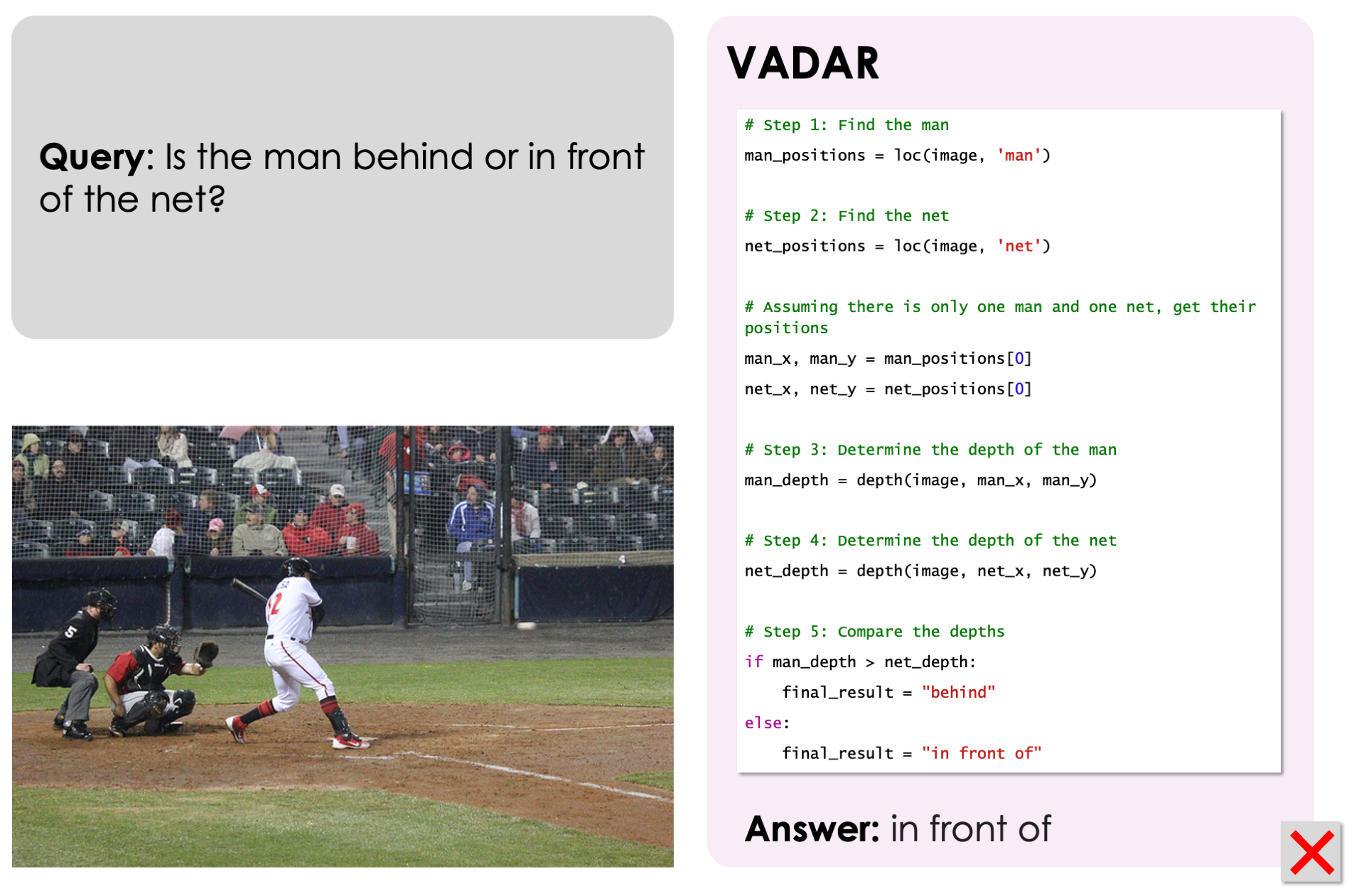}
    \vspace{-2mm}
    \caption{\method program outputs on GQA~\cite{gqa}.}
    \label{fig:gqa_examples}
    \vspace{-2mm}
\end{figure*}

\cref{fig:gqa_examples} shows qualitative examples on GQA~\cite{gqa}. Our method is able to identify and locate key objects necessary to answer questions. It is extremely explicit, locating the nearest person in the top right example using pixel distance from the tree. Some GQA questions have ambiguous answers, where the shape of the pot is generically ``round" and the frame of reference for spatial relations is not entirely clear (\ie, which man in the last example?).

\clearpage
\newtcblisting{psmall}{listing only,breakable, blank, borderline={1pt}{-5pt},listing options={breakindent=0pt,breaklines=true,basicstyle=\ttfamily\tiny}}

\begin{figure*}[t]
\begin{psmall}
\"\"\"
Locates objects in an image. Object prompts should be 1 WORD MAX.

Args:
    image (image): Image to search.
    object_prompt (string): Description of object to locate. Examples: "spheres", "objects".
Returns:
    list: A list of x,y coordinates for all of the objects located in pixel space.
\"\"\"
def loc(image, object_prompt):

\"\"\"
Answers a question about the attributes of an object specified by an x,y coordinate.
Should not be used for other kinds of questions.

Args:
    image (image): Image of the scene.
    question (string): Question about the objects attribute to answer. Examples: "What color is this?", "What material is this?"
    x (int): X coordinate of the object in pixel space.
    y (int): Y coordinate of the object in pixel space. 
    

Returns:
    string: Answer to the question about the object in the image.
\"\"\"
def vqa(image, question, x, y):

\"\"\"
Returns the depth of an object specified by an x,y coordinate.

Args:
    image (image): Image of the scene.
    x (int): X coordinate of the object in pixel space.
    y (int): Y coordinate of the object in pixel space.

Returns:
    float: The depth of the object specified by the coordinates.
\"\"\"
def depth(image, x, y):

\"\"\"
Checks if two pairs of coordinates correspond to the same object.

Args:
    image (image): Image of the scene.
    x_1 (int): X coordinate of object 1 in pixel space.
    y_1 (int): Y coordinate of object 1 in pixel space.
    x_2 (int): X coordinate of object 2 in pixel space.
    y_2 (int): Y coordinate of object 2 in pixel space.

Returns:
    bool: True if object 1 is the same object as object 2, False otherwise.
\"\"\"
def same_object(image, x_1, y_1, x_2, y_2):
\end{psmall}
\caption{\textbf{Pre-defined Modules for \clevr}. These modules are used to initialize the dynamic API. As \clevr defines size to be either \texttt{large} or \texttt{small}, we omit the \texttt{get\_2D\_object\_size} method.}
\label{fig:clevr_predefined}
\end{figure*}
\clearpage
\begin{figure*}[t]
\begin{psmall}
\"\"\"
Locates objects in an image. Object prompts should be 1 WORD MAX.

Args:
    image (image): Image to search.
    object_prompt (string): Description of object to locate.
Returns:
    list: A list of x,y coordinates for all of the objects located in pixel space.
\"\"\"
def loc(image, object_prompt):

\"\"\"
Answers a question about the attributes of an object specified by an x,y coordinate.
Should not be used for other kinds of questions.

Args:
    image (image): Image of the scene.
    question (string): Question about the objects attribute to answer. Examples: "What color is this?", "What material is this?"
    x (int): X coordinate of the object in pixel space.
    y (int): Y coordinate of the object in pixel space. 
    

Returns:
    string: Answer to the question about the object in the image.
\"\"\"
def vqa(image, question, x, y):

\"\"\"
Returns the depth of an object specified by an x,y coordinate.

Args:
    image (image): Image of the scene.
    x (int): X coordinate of the object in pixel space.
    y (int): Y coordinate of the object in pixel space.

Returns:
    float: The depth of the object specified by the coordinates.
\"\"\"
def depth(image, x, y):

\"\"\"
Checks if two pairs of coordinates correspond to the same object.

Args:
    image (image): Image of the scene.
    x_1 (int): X coordinate of object 1 in pixel space.
    y_1 (int): Y coordinate of object 1 in pixel space.
    x_2 (int): X coordinate of object 2 in pixel space.
    y_2 (int): Y coordinate of object 2 in pixel space.

Returns:
    bool: True if object 1 is the same object as object 2, False otherwise.
\"\"\"
def same_object(image, x_1, y_1, x_2, y_2):

\"\"\"
Returns the width and height of the object in 2D pixel space.

Args:
    image (image): Image of the scene.
    x (int): X coordinate of the object in pixel space.
    y (int): Y coordinate of the object in pixel space.

Returns:
    tuple: (width, height) of the object in 2D pixel space.
\"\"\"
def get_2D_object_size(image, x, y):
\end{psmall}
\caption{\textbf{Pre-defined Modules for \ourbench}. These modules are used to initialize the dynamic API.}
\label{fig:omni3d_predef}
\end{figure*}
\clearpage

\begin{figure*}[t]
\begin{psmall}
def loc(self, image, object_prompt):
    pts = molmo(image, "point to the " + object_prompt)
    if len(pts) == 0:
        # No points found
        return []
    return pts

def vqa(image, question, x, y):
    mask = sam_2([x, y], "foreground")  # get sam2 mask at x,y
    bbox = bbox_from_mask(mask)  # bbox around sam2 mask
    boxed_image = overlay_box_on_image(image, bbox)  # original image with bbox overlaid
    result = gpt4o(boxed_image, question)
    return result

def depth(image, x, y):
    depth_pred = unidepth(image)["depth"]  # Predict depth map over image
    depth_x_y = depth_pred[y, x]
    return depth_x_y

def same_object(image, x_1, y_1, x_2, y_2):
    mask_1 = sam_2([x_1, y_1], "foreground")  # get sam2 mask for point 1
    mask_2 = sam_2([x_2, y_2], "foreground")  # get sam2 mask for point 2
    obj_1_bbox = bbox_from_mask(mask_1)  # bbox around sam2 mask
    obj_2_bbox = bbox_from_mask(mask_2)  # bbox around sam2 mask
    return iou(obj_1_bbox, obj_2_bbox) > 0.92

def get_2D_object_size(image, x, y):
    mask = sam_2([x, y], "foreground")  # get sam2 mask at x,y
    bbox = bbox_from_mask(mask)  # bbox around sam2 mask
    width = abs(box[0] - box[2])
    height = abs(box[1] - box[3])
    return width, height
\end{psmall}
\caption{\textbf{Python Implementation of Predefined Modules.} \method uses Molmo~\cite{molmo} for object detection, SAM2~\cite{sam} for segmentation, GPT4o~\cite{gpt4} for VQA, and UniDepth~\cite{unidepth} for depth estimation.} 
\label{fig:predefined_implementation}
\end{figure*}
\clearpage
\newtcblisting{prompt}{listing only,breakable, blank, borderline={1pt}{-10pt},listing options={breakindent=0pt,breaklines=true,basicstyle=\ttfamily\scriptsize}}

\begin{figure*}[t]
\centering
\begin{prompt}
Propose only new method signatures to add to the existing API.

Available Primitives: image, int, string, list, tuple

Current API:
{current_api_signatures}

Next, I will ask you a series of questions that reference an image and are solvable with a python program that uses the API I have provided so far. Please propose new method signatures with associated docstrings to add to the API that would help modularize the programs that answer the questions. 

For each proposed method, output the docstring inside <docstring></docstring> immediately followed by the method signature for the docstring inside <signature></signature>. Do not propose methods that are already in the API.

Please ensure that you ONLY add new methods when necessary. Do not add new methods if you can solve the problem with combinations of the previous methods!

Added methods should be simple, building minorly on the methods that already exist.

Importantly, new methods MUST start with an underscore. As an example, you may define a "_get_material" method. Please ensure you ALWAYS start the name with an underscore.

Again, output the docstring inside <docstring></docstring> immediately followed by the method signature for the docstring inside <signature></signature>.

{questions}
\end{prompt}
\caption{\textbf{Signature Agent Prompt} used for both \clevr and \ourbench.}
\label{fig:signature_agent_prompt}
\end{figure*}
\begin{figure*}[t]
\begin{psmall}
Implement a method given a docstring and method signature, using the API specification as necessary.
Current API:
{pre_defined_signatures}
{generated_signatures}

Here are some examples of how to implement a method given its docstring and signature:
<docstring>
\"\"\"
Locates objects that are on the left of the reference object.
Args:
    image (IMAGE): Image to search.
    ref_x (int): X coordinate of reference object in pixel space.
    ref_y (int): Y coordinate of reference object in pixel space.
Returns:
    points (list): list of [x, y] coordinates for objects in pixel space matching description to the left.
\"\"\"
</docstring>
<signature>def objects_left(image, ref_x, ref_y):</signature>
<implementation>
objects_left = []
all_objects = loc(image, object_prompt='objects')
for object_point in all_objects:
    x, y = object_point
    if same_object(image, ref_x, ref_y, x, y):
        continue
    if x < ref_x:
        objects_left.append(object_point)
return objects_left
</implementation>
<docstring>
\"\"\"
Gets the material of the given object.
Args:
    image (IMAGE): Image that the object is contained in.
    ref_x (int): X coordinate of reference object in pixel space.
    ref_y (int): Y coordinate of reference object in pixel space.
Returns:
    str: Material of the object.
\"\"\"
</docstring>
<signature>def object_material(image, ref_x, ref_y):</signature>
<implementation>
material = vqa(image=image, question='What material is this object?', x=ref_x, y=ref_y)
return material
</implementation>
<docstring>
\"\"\"
Checks if an object 1 is in front of object 2.
Args:
    image (IMAGE): Image that the object is contained in.
    x_1 (int): X coordinate of object 1 in pixel space.
    y_1 (int): Y coordinate of object 1 in pixel space.
    x_2 (int): X coordinate of object 2 in pixel space.
    y_2 (int): Y coordinate of object 2 in pixel space.
Returns:
    bool: True if object 1 is in front of object 2, False otherwise
\"\"\"
</docstring>
<signature>def in_front_of(image, x_1, y_1, x_2, y_2):</signature>
<implementation>
depth_1 = depth(image, x_1, y_1)
depth_2 = depth(image, x_2, y_2)
return depth_1 < depth_2
</implementation>
<docstring>
\"\"\"
Checks if object1 has the same size as object2
Args:
    image (IMAGE): Image that the object is contained in.
    x_1 (int): X coordinate of object 1 in pixel space.
    y_1 (int): Y coordinate of object 1 in pixel space.
    x_2 (int): X coordinate of object 2 in pixel space.
    y_2 (int): Y coordinate of object 2 in pixel space.
Returns:
    bool: True if object 1 has the same size as object 2, False otherwise
\"\"\"
</docstring>
<signature>def same_size(image, x_1, y_1, x_2, y_2):</signature>
<implementation>
object_1_size = vqa(image=image, question='What size is this object?', x=x_1, y=y_1)
object_2_size = vqa(image=image, question='What size is this object?', x=x_2, y=y_2)
return object_1_size == object_2_size
</implementation>

Here are some helpful tips: 
1) When you need to search over objects satisfying a condition, remember to check all the objects that satisfy the condition and don't just return the first one. 
2) You already have an initialized variable named "image" - no need to initialize it yourself! 
3) When searching for objects to compare to a reference object, make sure to remove the reference object from the retrieved objects. You can check if two objects are the same with the same_object method.
Do not define new methods here, simply solve the problem using the existing methods.
Now, given the following docstring and signature, implement the method, using the API specification as necessary. Output the implementation inside <implementation></implementation>.
Again, Output the implementation inside <implementation></implementation>.
<docstring>{docstring}</docstring>
<signature>{signature}</signature>
\end{psmall}
\caption{\textbf{Implementation Agent Prompt for \clevr.} This prompt differs from the prompt used for \ourbench as we omit examples illustrating usage of the \texttt{get\_2D\_object\_size} method. The prompt features \emph{Weak ICL} examples illustrating correct usage of the pre-defined modules, as well as \emph{Pseudo ICL} in the form of natural language instructions.}
\label{fig:implementation_prompt_clevr}
\end{figure*}
\begin{figure*}[t]
\begin{psmall}
Implement a method given a docstring and method signature, using the API specification as necessary.
Current API:
{predef_signatures}
{generated_signatures}
Here are some examples of how to implement a method given its docstring and signature:
<docstring>
\"\"\" Locates objects that are on the left of the reference object.
Args:
    image (IMAGE): Image to search.
    ref_x (int): X coordinate of reference object in pixel space.
    ref_y (int): Y coordinate of reference object in pixel space.
Returns:
    points (list): list of [x, y] coordinates for objects in pixel space matching description to the left.
\"\"\"
</docstring>
<signature>def objects_left(image, ref_x, ref_y):</signature><implementation>
objects_left = []
all_objects = loc(image, object_prompt='objects')
for object_point in all_objects:
    x, y = object_point
    if same_object(image, ref_x, ref_y, x, y):
        continue
    if x < ref_x:
        objects_left.append(object_point)
return objects_left </implementation>
<docstring>
\"\"\" Gets the material of the given object.
Args:
    image (IMAGE): Image that the object is contained in.
    ref_x (int): X coordinate of reference object in pixel space.
    ref_y (int): Y coordinate of reference object in pixel space.
Returns:
    str: Material of the object.
\"\"\"
</docstring>
<signature>def object_material(image, ref_x, ref_y):</signature><implementation>
return vqa(image=image, question='What material is this object?', x=ref_x, y=ref_y) </implementation>
<docstring>
\"\"\" Checks if an object 1 is in front of object 2.
Args:
    image (IMAGE): Image that the object is contained in.
    x_1 (int): X coordinate of object 1 in pixel space.
    y_1 (int): Y coordinate of object 1 in pixel space.
    x_2 (int): X coordinate of object 2 in pixel space.
    y_2 (int): Y coordinate of object 2 in pixel space.
Returns:
    bool: True if object 1 is in front of object 2, False otherwise
\"\"\"
</docstring>
<signature>def in_front_of(image, x_1, y_1, x_2, y_2):</signature> <implementation>
depth_1, depth_2 = depth(image, x_1, y_1), depth(image, x_2, y_2)
return depth_1 < depth_2 </implementation>
<docstring>
\"\"\" Checks if object1 has the same size as object2
Args:
    image (IMAGE): Image that the object is contained in.
    x_1 (int): X coordinate of object 1 in pixel space.
    y_1 (int): Y coordinate of object 1 in pixel space.
    x_2 (int): X coordinate of object 2 in pixel space.
    y_2 (int): Y coordinate of object 2 in pixel space.
    epsilon (float): Acceptable margin of error in sizes.
Returns:
    bool: True if object 1 has the same size as object 2, False otherwise
\"\"\"
</docstring>
<signature>def same_size(image, x_1, y_1, x_2, y_2, epsilon):</signature> <implementation>
object_1_height, object_1_width = get_2D_object_size(image, x_1, y_1)
object_2_height, object_2_width = get_2D_object_size(image, x_2, y_2)
return abs(object_1_height - object_2_height) < epislon and abs(object_1_width - object_2_width) < epsilon </implementation>
<docstring>
\"\"\" Returns a list of objects in the images
Args:
    image (IMAGE): Image to search for objects in
Returns:
    list: List of strings corresponding to all of the objects in the image.
\"\"\"
</docstring>
<signature>def get_object_list(image):</signature> <implementation>
objects = []
object_points = loc(image, object_prompt='objects')
for object_point in object_coords:
    obj_x, obj_y = object_point
    objects.append(vqa(image, "What is this object?", obj_x, obj_y))
return objects </implementation> 
Here are some helpful definitions:
1) 2D distance/size refers to distance/size in pixel space. 2) 3D distance/size refers to distance/size in the real world. 3D size is equal to 2D size times the depth of the object. 3) "On" is defined as the closest object ABOVE another object. Only use this definition for "on". 4) "Next to" is defined as the closest object. 5) Width is the same as length. 6) "Depth" measures distance from the camera in 3D. 
Here are some helpful tips: 
1) When you need to search over objects satisfying a condition, remember to check all the objects that satisfy the condition and don't just return the first one. 2) You already have an initialized variable named "image" - no need to initialize it yourself! 3) When searching for objects to compare to a reference object, make sure to remove the reference object from the retrieved objects. You can check if two objects are the same with the same_object method. 4) Do not assume that the objects you see in these questions are all of the objects you will see, keep the methods general. 5) If two objects have the same 2D width, then the object with the largest depth has the largest 3D width. 6) If two objects have the same 2D height, then the object with the largest depth has the largest 3D height. 7) 2D sizes convey the height and width in IMAGE SPACE. To convert to height and width in 3D space, it needs to be multiplied by the depth! 8) If you are given a reference size, scale your output predicted size accordingly! Do not define new methods here, simply solve the problem using the existing methods. Now, given the following docstring and signature, implement the method, using the API specification as necessary. Output the implementation inside <implementation></implementation>. Again, Output the implementation inside <implementation></implementation>.
<docstring>
{docstring}
</docstring>
<signature>{signature}</signature>
\end{psmall}
\vspace{-3mm}
\caption{\textbf{Implementation Agent Prompt for \ourbench.} The prompt features \emph{Weak ICL} examples illustrating correct usage of the pre-defined modules, as well as \emph{Pseudo ICL} in the form of natural language instructions and definitions.}
\label{fig:implementation_prompt_omni3d}
\end{figure*}
\begin{figure*}[t]
\centering
\begin{prompt}
You are an expert logician capable of answering spatial reasoning problems with code. You excel at using a predefined API to break down a difficult question into simpler parts to write a program that answers spatial and complex reasoning problem.
Answer the following question using a program that utilizes the API to decompose more complicated tasks and solve the problem. 
Available sizes are {{small, large}}, available shapes are {{square, sphere, cylinder}}, available material types are {{rubber, metal}}, available colors are {{gray, blue, brown, yellow, red, green, purple, cyan}}.
The question may feature attributes that are outside of the available ones I specified above. If that's the case, please replace them to the most appropriate one from the attributes above.
I am going to give you an example of how you might approach a problem in psuedocode, then I will give you an API and some instructions for you to answer in real code.

Example:
Question: "What is the shape of the matte object in front of the red cylinder?"
Solution:
1) Find all the cylinders (loc(image, 'cylinders'))
2) If cylinders are found, loop through each of the cylinders found
3) For each cylinder found, check if the color of this cylinder is red. Store the red cylinder if you find it and break from the loop.
4) Find all the objects.
5) For each object, check if the object is rubber (matte is not in the available attributes, so we replace it with rubber)
6) For each rubber object O you found, check if the depth of O is less than the depth of the red cylinder
7) If that is true, return the shape of that object

Now here is an API of methods, you will want to solve the problem in a logical and sequential manner as I showed you
------------------ API ------------------
{pre_defined_signatures}
{api}
------------------ API ------------------
Please do not use synonyms, even if they are present in the question.
Using the provided API, output a program inside the tags <program></program> to answer the question. 
It is critical that the final answer is stored in a variable called "final_result".
Ensure that the answer is either yes/no, one word, or one number.
Here are some helpful tips: 
1) When you need to search over objects satisfying a condition, remember to check all the objects that satisfy the condition and don't just return the first one. 
2) You already have an initialized variable named "image" - no need to initialize it yourself! 3) Do not define new methods here, simply solve the problem using the existing methods.
3) When searching for objects to compare to a reference object, make sure to remove the reference object from the retrieved objects. You can check if two objects are the same with the same_object method.
Again, available sizes are {{small, large}}, available shapes are {{square, sphere, cylinder}}, available material types are {{rubber, metal}}, available colors are {{gray, blue, brown, yellow, red, green, purple, cyan}}.
Again, answer the question by using the provided API to write a program in the tags <program></program> and ensure the program stores the answer in a variable called "final_result".
It is critical that the final answer is stored in a variable called "final_result".
Ensure that the answer is either yes/no, one word, or one number.
AGAIN, answer the question by using the provided API to write a program in the tags <program></program> and ensure the program stores the answer in a variable called "final_result".
You do not need to define a function to answer the question - just write your program in the tags. Assume "image" has already been initialized - do not modify it!
<question>{question}</question>
\end{prompt}
\caption{\textbf{Program Agent Prompt for \clevr.} In the prompt, we provide a list of all available attributes in \clevr, a \emph{Pseudo ICL} example in natural language, and some helpful tips.}
\label{fig:program_prompt_clevr}

\end{figure*}
\begin{figure*}[t]
\centering
\begin{prompt}
You are an expert logician capable of answering spatial reasoning problems with code. You excel at using a predefined API to break down a difficult question into simpler parts to write a program that answers spatial and complex reasoning problem.
Answer the following question using a program that utilizes the API to decompose more complicated tasks and solve the problem. 
I am going to give you two examples of how you might approach a problem in psuedocode, then I will give you an API and some instructions for you to answer in real code.

Example 1:
Question: "What is the shape of the red object in front of the blue pillow?"
Solution:
1) Find all the pillows (loc(image, 'pillow')).
2) If pillows are found, loop through each of the pillows found.
3) For each pillow found, check if the color of this pillow is blue. Store the blue pillow if you find it and break from the loop.
4) Find all the objects.
5) For each object, check if the object is red.
6) For each red object O you found, check if the depth of O is less than the depth of the blue pillow.
7) If that is true, return the shape of that object.

Example 2:
Question: "How many objects have the same color as the metal bowl?"
Solution:
1) Set a counter to 0
2) Find all the bowls (loc(image, 'bowls')).
3) If bowls are found, loop through each of the bowls found.
4) For each bowl found, check if the material of this bowl is metal. Store the metal bowl if you find it and break from the loop.
5) Find and store the color of the metal bowl.
6) Find all the objects.
7) For each object O, check if O is the same object as the small bowl (same_object(image, metal_bowl_x, metal_bowl_y, object_x, object_y)). If it is, skip it.
8) For each O you don't skip, check if the color of O is the same as the color of the metal bowl.
9) If it is, increment the counter.
10) When you are done looping, return the counter.

Now here is an API of methods, you will want to solve the problem in a logical and sequential manner as I showed you
------------------ API ------------------
{predef_signatures}
{api}
------------------ API ------------------
Please do not use synonyms, even if they are present in the question.
Using the provided API, output a program inside the tags <program></program> to answer the question. 
It is critical that the final answer is stored in a variable called "final_result".
Ensure that the answer is either yes/no, one word, or one number.
Here are some helpful definitions:
1) 2D distance/size refers to distance/size in pixel space.
2) 3D distance/size refers to distance/size in the real world. 3D size is equal to 2D size times the depth of the object.
3) "On" is defined as the closest object ABOVE another object. Only use this definition for "on".
4) "Next to" is defined as the closest object.
5) Width is the same as length.
6) "Depth" measures distance from the camera in 3D.
Here are some helpful tips: 
1) When you need to search over objects satisfying a condition, remember to check all the objects that satisfy the condition and don't just return the first one. 
2) You already have an initialized variable named "image" - no need to initialize it yourself! 
3) When searching for objects to compare to a reference object, make sure to remove the reference object from the retrieved objects. You can check if two objects are the same with the same_object method.
4) Do not assume that the objects you see in these questions rae all of the objects you will see, keep the methods general.
5) If two objects have the same 2D width, then the object with the largest depth has the largest 3D width.
6) If two objects have the same 2D height, then the object with the largest depth has the largest 3D height.
7) 2D sizes convey the height and width in IMAGE SPACE. To convert to height and width in 3D space, it needs to be multiplied by the depth!
8) If you are given a reference size, scale your output predicted size accordingly!
Again, answer the question by using the provided API to write a program in the tags <program></program> and ensure the program stores the answer in a variable called "final_result".
It is critical that the final answer is stored in a variable called "final_result".
Ensure that the answer is either yes/no, one word, or one number.
AGAIN, answer the question by using the provided API to write a program in the tags <program></program> and ensure the program stores the answer in a variable called "final_result".
You do not need to define a function to answer the question - just write your program in the tags. Assume "image" has already been initialized - do not modify it!
<question>{question}</question>
\end{prompt}
\caption{\textbf{Program Agent Prompt for \ourbench.} The prompt features \emph{Pseudo ICL} in the form of two natural language examples and helpful tips for handling 2D and 3D dimensions.}
\label{fig:program_prompt_omni3d}
\end{figure*}

\end{document}